%% file: main.tex
\documentclass[10pt,twocolumn,letterpaper]{article}

\usepackage[pagenumbers]{cvpr} % To force page numbers, e.g. for an arXiv version

\input{preamble}

\definecolor{cvprblue}{rgb}{0.21,0.49,0.74}
\usepackage[pagebackref,breaklinks,colorlinks,allcolors=cvprblue]{hyperref}

\def\confName{CVPR}
\def\confYear{2026}

\title{Inferring Dynamic Physical Properties from Video Foundation Models}

\author{%
  Guanqi Zhan$^{1*}$, Xianzheng Ma$^{1*}$,
  Weidi Xie$^{1,2}$, Andrew Zisserman$^1$\\
    $^1$VGG, University of Oxford\quad\quad
    $^2$Shanghai Jiao Tong University\\
  \texttt{\{guanqi,xianzheng,weidi,az\}@robots.ox.ac.uk} 
}

\usepackage{algorithm}
\usepackage{algorithmic}
\usepackage{booktabs}
\usepackage{amssymb}
\usepackage{comment}
\usepackage{multirow}
\usepackage{amsmath}
\usepackage{graphicx}

\begin{document}
\maketitle

\def\thefootnote{*}\footnotetext{Equal contribution.}\def\thefootnote{\arabic{footnote}}

\input{sec/0_abstract}    
\input{sec/1_intro}

\input{sec/2_related}
\input{sec/3_preliminary}

\input{sec/3_data}

\input{sec/4_architecture}
\input{sec/5_experiment}

\input{sec/6_conclusion}

\vspace{2pt} 
\noindent \textbf{Acknowledgements.} This research is supported by EPSRC Programme Grant VisualAI EP$\slash$T028572$\slash$1, a Royal Society Research Professorship RP$\backslash$R1$\backslash$191132 and a China Oxford Scholarship.
We thank Minghao Chen, Shuai Chen, Jindong Gu, João Henriques, Zeren Jiang, Mengmeng Li, Zixuan Li, Shuai Mao, Boyu Pang, Ashish Thandavan, Jianyuan Wang, Junyu Xie, Wen Xiong and Chuanxia Zheng for their help and support for the project.

{
    \small
    \bibliographystyle{ieeenat_fullname}
    \bibliography{main}
}

\appendix
\input{sec/7_supple}

\end{document}

%% file: preamble.tex
\newcommand{\datasetname}{\textit{PhysVid}}

%% file: sec/0_abstract.tex
\begin{abstract}

We study the task of predicting dynamic physical properties from videos. More specifically, we consider physical properties that require {\em temporal} information to be inferred: elasticity of a bouncing object, viscosity of a flowing liquid, and dynamic friction of an object sliding on a surface. To this end, we make the following contributions: (i) We collect a new video dataset for each physical property, consisting of synthetic training and testing splits, as well as a real split for real world evaluation. (ii) We explore three ways to infer the physical property from videos: (a) an oracle method where we supply the visual cues that intrinsically reflect the property using classical computer vision techniques; (b) a simple read out mechanism using a visual prompt and trainable prompt vector for cross-attention on pre-trained video generative and self-supervised models; and (c) prompt strategies for Multi-modal Large Language Models (MLLMs). (iii) We show that a video foundation model trained in a generative (DynamiCrafter) or trained in a self-supervised manner (V-JEPA-2) achieve a generally similar performance, though behind that of the oracle, and that MLLMs are currently inferior to the other models, though their performance can be improved through suitable prompting. The dataset, model, and code are available at \url{https://www.robots.ox.ac.uk/~vgg/research/idpp/}.

\end{abstract}

%% file: sec/1_intro.tex
\section{Introduction}

Humans are remarkably adept at intuitively estimating physical properties from visual observations. Without direct interaction, people can often estimate how bouncy a ball is, how thick a liquid seems, or how slippery a surface might be—simply by watching how objects move. While these estimations are not precise in a scientific sense, they are sufficiently accurate for guiding perception, prediction, and action. 
Bringing this capability to machines is an important step towards building more general and physically grounded artificial intelligence. In particular, visual systems that can infer dynamic physical properties from raw video could enhance robotic manipulation, embodied agents, and video understanding tasks in ways that go beyond the traditional perception tasks of recognition, detection, and segmentation.

Recent progress in video foundation models, including generative models~\citep{xing2024dynamicrafter, liu2024sora}, self-supervised models~\citep{bardes2023vjepa, assran2025vjepa2} and multi-modal large language models (MLLMs)~\citep{hui2024qwen2,comanici2025gemini,hurst2024gpt}, have shown impressive capability in synthesizing realistic dynamics, learning general-purpose video representations, and tackling semantic understanding tasks, for example, video question answering. 
However, a question that remains underexplored is: \textbf{do these models acquire an understanding of {\em dynamic} physical properties from videos?}

In this paper, we address this question by focusing on several representative physical properties that are not directly observable in static frames but instead emerge through temporal dynamics: the elasticity of a bouncing object, the viscosity of a flowing liquid, and the dynamic friction between a surface and a sliding object. These properties are especially compelling because their inference requires temporal reasoning and sensitivity to subtle visual cues—such as deformation, deceleration, spreading, or oscillation. By examining how well current video foundation models capture these dynamic attributes, we aim to assess their physical understanding beyond static appearance. 

To support this investigation, we introduce a new dataset, \datasetname, specifically designed to evaluate the dynamic physical properties from video. Existing datasets lack ground-truth annotations for such properties, so we construct \datasetname~using a combination of synthetic videos—rendered via a physics simulator—and real-world videos sourced from the internet or captured in-house. Each video is annotated with physical property values, either derived from simulation parameters or estimated manually. 
The dataset is designed to facilitate the study of out-of-domain generalization, both within the synthetic domain and from synthetic to real-world data. 
To establish an upper bound on what is inferable from visual input alone, we implement an oracle method for each property. These oracles leverage privileged access to the visual cues that directly reflect the corresponding property.

We evaluate three categories of video foundation models: generative models, self-supervised models, and multi-modal large language models (MLLMs). For the generative and self-supervised models, we propose a simple yet effective readout mechanism that extracts dynamic physical properties from pre-trained, frozen representations. Our method introduces a learnable query vector that attends to internal representation tokens via cross-attention, enabling the selective extraction of relevant information. 
This approach is both lightweight and training-efficient. More specifically, we have studied DynamiCrafter~\cite{xing2024dynamicrafter} for a video generative model, and V-JEPA-2~\cite{assran2025vjepa2} for a video self-supervised model. For MLLMs, we explore various prompting strategies to elicit predictions of dynamic physical properties directly from video input. These strategies include few-shot prompting to provide task context, as well as procedural prompting that guides the model through the oracle estimation steps—helping it focus on the intrinsic visual cues that reveal the target properties. The MLLMs we have studied include QwenVL\citep{hui2024qwen2}, GPT~\citep{hurst2024gpt}, and Gemini~\citep{comanici2025gemini}.

%% file: sec/2_related.tex
\section{Related Work}

\noindent \textbf{Physics Prediction from Images and Videos.}
Inferring physical properties from visual observations remains a core challenge in computer vision.
Early methods estimate latent physical parameters (e.g., mass, friction, stiffness) via differentiable physics engines or learning-based simulators~\citep{wu2015galileo,ding2021dynamic,gradsim,LiLinYiBeaYam20,KunMriKos20,WanKryDenAscHua18}, while later works infer salient attributes like viscosity or elasticity from task-specific visual cues~\citep{TakKazRolShi14,PaukawNisShiFle15,AssBarFle18,NorWieNorTayCra07,TakShi16,PauSchFilAssFle17,PauFle20}, yet both rely heavily on simulation supervision, domain priors, or handcrafted heuristics.
More recently, unsupervised learning of intuitive physics has emerged via next-frame prediction from large-scale everyday physical scenes~\citep{voleti2022masked,lu2023vdt,AgrNaiAshAbbMal16,CheSer17,fitvid,hafner2019learning,billiard,garcia2025learning}, capturing latent dynamics without explicit physical supervision. However, the resulting representations are usually implicit and lack interpretability in terms of concrete physical quantities.
In contrast, we infer physical properties by directly prompting pre-trained video foundation models, enabling explicit estimation without reliance on task-specific heuristics, or end-to-end prediction pipelines from scratch.

\noindent \textbf{Physics Datasets and Benchmarks.} 
An increasing number of physics-related datasets have been collected in recent years to provide ground truth annotations for different physical properties, including material~\citep{sharma2023materialistic,gao2024physically}, shadow~\citep{Wang_2020_soba,Wang_2021_soba}, support relations~\citep{silberman2012indoor},occlusion~\citep{zhan2022triocc,zhan2024amodal}, mass and volume~\citep{wu2016physics101}. 
Another line of work~\citep{chow2025physbench,shen2025phyx,riochet2018intphys,bordes2025intphys2,tung2023physion++,bear2021physion} proposes broad benchmarks with video-image-text QA tasks to assess physical understanding in vision-language models, but the questions are typically qualitative and categorical. 
More recently, ~\citet{zhang2025morpheus} introduces a benchmark consisting of 130 real-world videos capturing physical phenomena guided by conservation laws, to evaluate the physics plausibility of video generative models by assessing the trajectory of objects in their generated videos.
In contrast, our datasets consist of both \emph{synthetic} and \emph{real-world} videos annotated with the \emph{quantitative value} for the associated physical parameter of the coefficient of friction, elasticity, and viscosity.

%% file: sec/3_preliminary.tex
\section{Problem Scenario and The \datasetname~Datasets}
\label{sec:problem}

In this paper, we address the problem of estimating physical properties from videos. Specifically, we focus on three properties: 
\textbf{elasticity} of a bouncing object, 
\textbf{viscosity} of a flowing liquid, and the \textbf{dynamic friction coefficient} between a surface and a sliding object. 
Given a video $v \in \mathbb{R}^{T \times H \times W \times 3}$, we consider two formulations, the first is \textbf{absolute value prediction}, where the input is a single video and the model is tasked with predicting the numerical value of the physical property,
{\em i.e.}, $y_{\text{abs}} = \Phi(v;\theta_1)$.
The second is \textbf{relative value comparison}, 
where the input is a pair of videos captured from the same viewpoint, 
and the model must determine whether the first video exhibits a higher physical property value than the second, 
{\em i.e.}, $y_{\text{rel}} = \Phi(v_1, v_2;\theta_2)$, 
and $y_{\text{rel}}$ is binary.

Each scenario is parameterized by a set of variables, including the value of the target {\em physical property}~(\emph{e.g.,} elasticity, viscosity, or friction), and a set of {\em nuisance parameters}~(including camera viewpoint, object appearance, lighting, {\em etc.}). While the model must be sensitive to changes in the physical property, it should be robust (ideally invariant) to variations in nuisance parameters.

To assess generalization, we define two domains of nuisance parameters, denoted as $\mathcal{A}_1$ and $\mathcal{A}_2$, which differ in their distributions. For instance, $\mathcal{A}_2$ may have different camera viewpoints or different lighting conditions to $\mathcal{A}_1$ (full details of these differences are given in Appendix Section~\ref{sec:supple_dataset_details}).
We generate a dataset using a physics-based simulator, consisting of one training split and two test splits. 
The models are only trained on the training split  from the simulator for all the evaluations.
The training and \texttt{test-1} splits are sampled from $\mathcal{A}_1$, sharing the same nuisance distribution; \texttt{test-2} is drawn from $\mathcal{A}_2$, introducing a distribution shift. 
The target property values are sampled from a shared range across all splits to ensure consistency.
Finally, \texttt{test-3} consists of real-world videos, 
used to evaluate generalization beyond simulation.

%% file: sec/3_data.tex
\subsection{The \datasetname~Datasets}
\label{sec:data}

\begin{figure*}[t]
	\centering
\includegraphics[width=\textwidth]{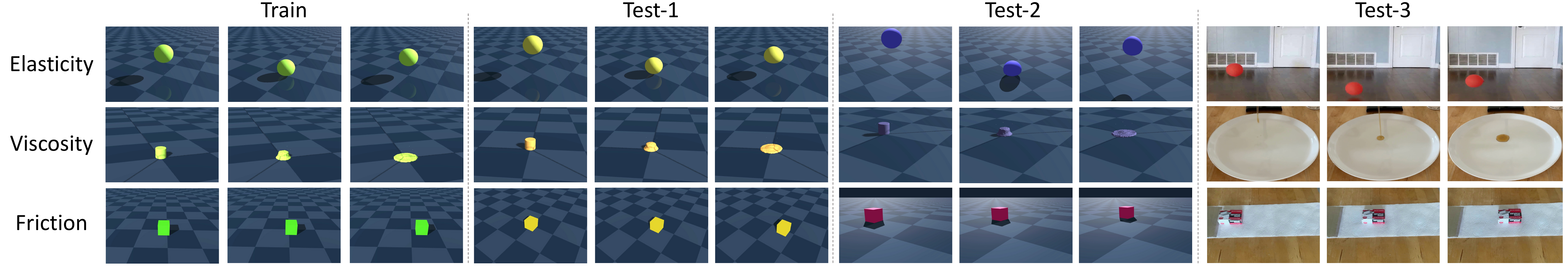}
\caption{
\textbf{Examples of the \datasetname~dataset.} 
Each row shows a different property, and each column shows three frames from video samples in the synthetic sets (\texttt{train}, \texttt{test-1}, and \texttt{test-2}) and the real \texttt{test-3} set. The \texttt{train} and \texttt{test-1} sets are from the same distribution. In \texttt{test-2} parameters, such as lighting, viewpoint and color, differ from those in \texttt{test-1}. 
}
\label{fig:data_sample}
\end{figure*}

To study the dynamic physical properties of elasticity, viscosity, and friction, we construct a dataset for each, containing both synthetic and real-world videos. Synthetic ones are generated with the Genesis simulator~\citep{Genesis}, and real ones are captured with an iPhone in slow-motion mode or downloaded from the Internet.
For each property we have: $10,000$ videos for \texttt{train}; $1000$ videos for each of 
\texttt{test-1} and \texttt{test-2}; and $100$ videos for \texttt{test-3}.
Sample frames are shown in Figure~\ref{fig:data_sample}.
In the following we describe how each property is realized in the video. Please refer to Appendix Section~\ref{sec:supple_dataset_details} for more details of the datasets.

\subsubsection{Elasticity}

We study an object’s elasticity by analyzing the motion of a ball dropped onto the ground and its subsequent bounces. In physics, elasticity $e$ is quantified as the ratio of the rebound velocity $v_{\text{after impact}}$ to the impact velocity $v_{\text{before impact}}$, and also equals $\sqrt{h_{\text{bounce}} / h_{\text{drop}}}$, where $h_{\text{drop}}$ is the dropping height and $h_{\text{bounce}}$ is the bouncing height. Here and for the following properties, please refer to Appendix Section~\ref{sec:derivation_physics} for the detailed derivations. These expressions are used for the  oracle estimation in Section~\ref{sec:arch_oracle}.

\noindent \textbf{Synthetic Dataset.} 
All synthetic videos are generated using Genesis~\citep{Genesis}, 
with object’s elasticity as the target property. 
Nuisance factors include drop height, camera viewpoint, object appearance, and lighting conditions. The object is of the same size in all videos. Note, here and for the following properties, the ground truth property value is obtained directly from the simulator.

\noindent \textbf{Real-World Dataset.} 
The real-world videos are collected from YouTube using the search term ``ball bouncing experiments''. Each clip is manually trimmed to include the drop-and-bounce sequence of a single ball. The dataset includes a wide range of materials~({\em e.g.}, rubber balls, tennis balls, basketballs, balloons, {\em etc}), resulting in diverse elasticity values. 
The ground truth elasticity values for the real sequences are estimated by computing $\sqrt{h_{\text{bounce}} / h_{\text{drop}}}$: the videos are chosen such that the balls bounce in a fronto-parallel plane, which means that ratios of image heights (differences in $y$-coordinates) are approximately equal to the ratio of heights in 3D. These image differences are obtained by manual annotation.

\subsubsection{Viscosity}

We study the viscosity by observing a liquid column dropping and spreading on the ground. The viscosity can be reflected by the growth rate of the liquid area on the ground. The viscosity $\mu$ is negatively correlated to the liquid area growth rate $\frac{d(A(t))}{dt}$, given the controlled liquid density $\rho$, controlled liquid column diameter $D$, and controlled dropping velocity $v$ of the liquid column when it reaches the ground.

\noindent \textbf{Synthetic Dataset.} The synthetic videos are generated using Genesis~\citep{Genesis}, where the target property is the viscosity of liquid. Nuisance factors include camera viewpoint, object appearance, and lighting conditions. The liquid column is of the same size in all videos.

\noindent \textbf{Real-World Dataset.} 
Since it is challenging to find real-world videos online that provide ground-truth viscosity values while controlling for other relevant physical parameters—such as $\rho$, $D$ and $v$, we collected real videos under controlled conditions. We use a funnel with a fixed nozzle diameter to produce a consistent liquid column. A funnel holder allows us to fix the height from which the liquid is poured, thereby controlling the initial velocity $v$.
Ground-truth viscosity values for each liquid are obtained from standard physics reference tables. The selected liquids span a wide range of viscosities, from 1.2 ({\em e.g.}, coffee) to 225 ({\em e.g.}, maple syrup), allowing for a diverse and comprehensive evaluation.

\subsubsection{Friction}

We study friction between an object and a surface by observing how the object slows down as it slides with an initial velocity. The dynamic friction coefficient $\mu_k$ is proportional to the (negative) acceleration of the object $a$.

\noindent \textbf{Synthetic Dataset.} 
The synthetic videos are generated using Genesis~\citep{Genesis}, where the target property is the dynamic friction coefficient at the contacting surface of the object and the ground. Nuisance factors include initial location and initial velocity of the object, camera viewpoint, object appearance, and lighting conditions. The object is of the same size in all videos.

\noindent \textbf{Real-World Dataset.} 
While many online videos depict objects sliding on surfaces, 
they lack ground-truth annotations for friction coefficients. 
We therefore collect a real video dataset featuring 5 different objects and 6 surface materials, spanning a wide range of dynamic friction values. Each object is given an initial velocity by sliding it down from a slope and it then slides on a horizontal plane. 
To obtain ground-truth friction coefficients, we use a spring dynamometer to measure the friction force $F$ for each object-surface pair (by dragging the object at constant speed), and record the object’s weight $G$. The dynamic friction coefficient is then computed as: $\mu_k = F / G$.

%% file: sec/4_architecture.tex
\section{Inferring Physical Properties}
\label{sec:arch_visual}

This section presents the three different ways for inferring dynamic physical properties: an oracle method via classical computer vision techniques~(Section~\ref{sec:arch_oracle}); a visual prompt mechanism for video generative and self-supervised models~(Section~\ref{sec:arch_generative_self_supervised}); and prompts for MLLMs~(Section~\ref{sec:arch_language}).

\begin{figure*}[t]
	\centering
\includegraphics[width=\textwidth]{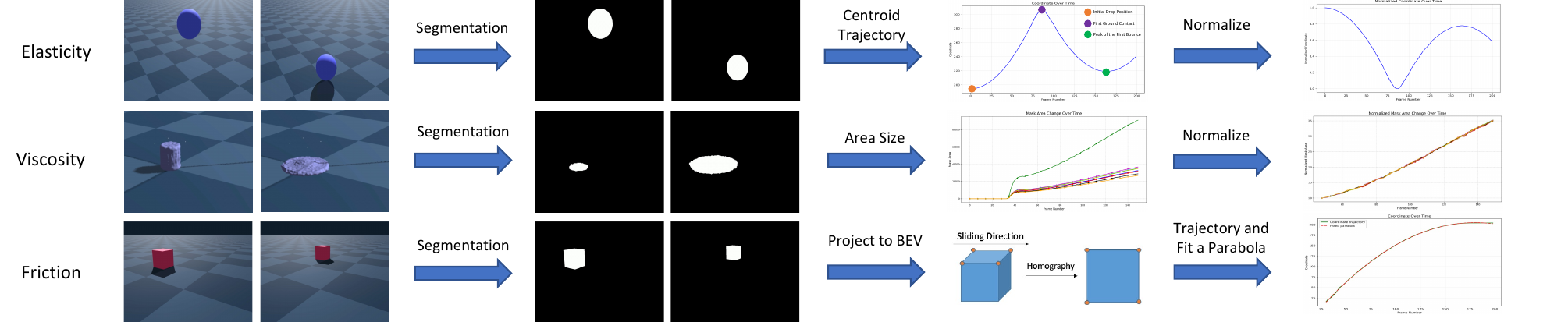}
\caption{
\textbf{Oracle methods for physical properties.} 
The objective in each case is to extract a measurement from the sequence that can directly be used to predict the property. For elasticity, we extract the centroid trajectory from segmentation masks, and then normalize the $y$-coordinates into $0$-$1$; the ratio of bouncing to dropping height over the sequence indicates the elasticity. For viscosity, we calculate the area size in the image via segmentation masks, and then normalize the area sizes by the area in the frame when the liquid first touches the ground; the slope of the normalized area size sequence reflects the viscosity. For friction, we transform to a bird's eye view (using a  homography transformation based on 4 corner points of the top surface of the sliding object), and fit a parabola $x = \alpha t^2 + \beta t + c$  
to the transformed trajectory; the parabola coefficient $\alpha$ predicts the friction coefficient.
For each video, we show the segmentation for two frames (left $\rightarrow$ right). 
}
\label{fig:oracle_pipe}
\end{figure*}

\subsection{Oracle Estimation}
\label{sec:arch_oracle}

\subsubsection{Elasticity} 
We aim to estimate elasticity from both synthetic and real-world videos. The key visual cue is the relative height of the ball during its drop and subsequent bounce, observed in 3D. As noted earlier, 
the ratio in 3D can be approximated from their corresponding image-space measurements. This approximation is exact when the motion occurs in a fronto-parallel plane, and remains reasonably accurate otherwise—since the ratio of lengths between parallel line segments is invariant under affine transformations~\citep{Hartley04c}. Given that perspective effects are minimal in our videos, the affine approximation provides a reliable estimate for elasticity. 
To automate this process, we extract the ball’s trajectory $y(t)$ 
from the video and input the sequence of ratios into a GRU network to regress the elasticity. In detail, we segment the ball in each frame and use their centroids as the $y$-coordinate. 
From this trajectory, we identify key points: the initial drop position, the first ground contact, and the peak of the first bounce. The resulting trajectory is normalized to the range $[0,1]$, by subtracting the $y$-coordinate of the first ground contact and dividing by the initial drop height. This normalization not only ensures invariance to viewpoint and scale, but also simplifies learning for the GRU by standardizing the input distribution. We train a GRU, as it is noisy to directly obtain $h_{\text{drop}}$ and $h_{\text{bounce}}$ using heuristics (\emph{e.g.,} determining the maximum and minimum points), and in practice a GRU provides a good estimate. 
The full pipeline is illustrated in Figure~\ref{fig:oracle_pipe} (top row). For the \textbf{absolute prediction}, the normalized trajectory is fed into a GRU network, which directly regresses the elasticity value. For the \textbf{relative comparison}, the binary decision score between two videos $v_1$ and $v_2$ is calculated as: 
\begin{equation}
\label{eq:relative_score}
\texttt{score} = \sigma(\log (\frac{e_1}{e_2})),
\end{equation}
where $e_1$ and $e_2$ are the estimated elasticities based on height ratios, and $\sigma(\cdot)$ denotes the sigmoid function.

\subsubsection{Viscosity} 
The key visual cue for estimating viscosity is the rate at which the liquid spreads on the ground-plane, measured as an area ratio normalized by the initial area of the liquid column. As with elasticity, we approximate perspective using an affine transformation -- here of the ground-plane. Since area ratios are invariant under affine transformations~\citep{Hartley04c}, the liquid’s normalized image-space area growth approximates its true normalized ground-plane expansion (in our setup the liquid spreads only within a limited area around the release point, and the camera is distant; consequently an affine viewing approximation is adequate).
Specifically, we extract segmentation masks for each frame and compute the liquid’s area over time. This area sequence is normalized by the area in the first frame where the liquid contacts the surface, ensuring invariance to viewpoint and scale. The process is illustrated in Figure~\ref{fig:oracle_pipe} (middle row).
For \textbf{absolute prediction}, we calculate the slope $k$ of $A(t)$ and use $1/k$ to represent the viscosity value; 
For \textbf{relative comparison}, the binary decision score between two videos $v_1$ and $v_2$ is calculated as in Equation~\ref{eq:relative_score}, where $e_1$ and $e_2$ are the estimated viscosities based on area growth rate.

\subsubsection{Friction} 
The key visual cue for estimating dynamic friction is the acceleration of the sliding object—{\em i.e.}, how quickly its velocity decreases due to friction—which can be inferred from its position over time. Since the object moves significantly in the video, we do not use an affine approximation, but instead take account of the projective geometry by 
mapping the object's motion to a bird’s-eye view, allowing for consistent trajectory analysis.
This is achieved by estimating a homography between the image and bird's eye view (normal to the plane) from the four corners of the object's top surface (see Figure~\ref{fig:oracle_pipe}, bottom row). 
We fit a parabola $x = \alpha t^2 + \beta t + c $ to the transformed top surface trajectory to estimate the acceleration $a$ from the coefficient $\alpha$, and the coefficient of friction $ \mu_k = 2 \alpha / g$. 
For \textbf{absolute prediction}, 
we use the estimated $\mu_k$
to represent the friction coefficient value; 
For \textbf{relative comparison}, the binary decision score between two videos $v_1$ and $v_2$ is calculated as in Equation~\ref{eq:relative_score}, where $e_1$ and $e_2$ are the estimated friction coefficients based on the transformed object trajectory.

\begin{figure*}[t]
	\centering
\includegraphics[width=\textwidth]{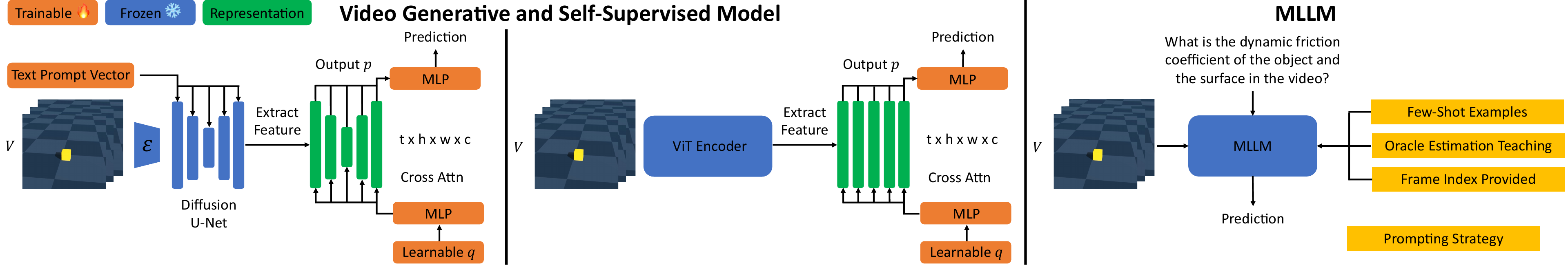}
\caption{
\textbf{Architectures for dynamic physical property prediction.} 
\textbf{Left}: video generative model as backbone; 
\textbf{Middle}: video self-supervised model as backbone; 
\textbf{Right}: multimodal large language model~(MLLM).
For the pre-trained video diffusion model (U-Net, left) and the pre-trained self-supervised model (ViT, middle), the representations are kept frozen, and a `visual prompt' learns to infer the physical properties.
For the MLLMs, the physical properties are inferred using a language prompt (right). 
} 
\label{fig:arch}
\end{figure*}

\subsection{Video Generative and Self-Supervised Models}
\label{sec:arch_generative_self_supervised}

\subsubsection{Video Feature Extraction}
\label{sec:feature_extraction}

Given a video $v \in \mathbb{R}^{T \times H \times W \times 3}$, 
we extract features with a pre-trained video backbone,
that can either be generative or self-supervised, 
resulting into spatiotemporal feature representations, 
{\em i.e.}, $r = \psi(v) \in \mathbb{R}^{t \times h \times w \times c}$, which can be detailed as follows.

\noindent \textbf{Generative Model as Backbone.}
We adopt a pre-trained video diffusion model~(Figure~\ref{fig:arch}, left), namely DynamiCrafter~\citep{xing2024dynamicrafter}, 
to compute the visual features. 
Specifically, given an input video, we add noise to the latent representations after the pre-trained VAE encoder, and replace the text prompt with a learnable embedding. We extract multi-scale features from all U-Net layers at diffusion time step 50, 
which was shown to be effective for capturing 3D physics in prior work~\citep{tang2023emergent, zhan2024general}. 
To aggregate the features, we introduce a learnable query vector $q$, which is first mapped to the different dimensions of the multi-scale features (see Appendix Section~\ref{sec:supple_imple_detail_dimension_mlp} for details), and then attends to the diffusion tokens~($r_i$) via cross-attention:
\begin{equation}
    p = \sum_{i=1}^{t \times h \times w}\texttt{softmax}(q \cdot r_i) \cdot r_i
\end{equation}
The resulting vectors $p$ from different layers are then mapped by another MLP network to a common  dimension and average pooled to generate the final video feature representation $P$.
To predict the physical properties, we train the text token of the generative model, together with the `visual prompt' architecture that includes the query $q$ and the MLPs.

\noindent \textbf{Self-Supervised Model as Backbone.}
Here, we adopt a pre-trained self-supervised model~(Figure~\ref{fig:arch}, middle), namely V-JEPA-2~\citep{assran2025vjepa2}, as the visual backbone.
The input video is passed through the model, and we extract feature tokens from all layers of the ViT encoder. Similar to the generative setting, we introduce a learnable query vector $q$ to extract the video feature representation $P$ from the ViT tokens via attentive pooling.
Although the feature dimension at each ViT layer is the same, we still use a MLP network to map $q$ to generate the query vector of each layer (keeping it similar to the generative setting in terms of MLP network architecture), and use another MLP network to map the output vectors $p$ to a same dimension as the generative setting before average pooling them to get $P$. Please see Appendix Section~\ref{sec:supple_imple_detail_dimension_mlp} for more details.

\subsubsection{Physical Property Prediction}
\label{sec:property_prediction}

Given the computed feature $P$ from video foundation models, 
we train a MLP network to predict the physical properties using the synthetic video dataset training split. The network for each property is trained separately.

\noindent \textbf{Absolute Value Prediction.}
Given the resulting video feature~($P$), we pass it through a MLP network $\gamma$ to predict the absolute value $\chi$ of the physical property:
\begin{equation}
    \chi = \gamma (P)
\end{equation}
For elasticity and friction, the absolute value prediction is supervised with L1 loss with the ground truth value; For viscosity, as the ground truth values may have very different scales, \emph{i.e.,} from $1e^{-5}$ to $1e^{-2}$, the absolute value prediction is trained with Log L1 loss, which calculates L1 loss between the log of the predicted value and the log of the ground truth value. 

\noindent \textbf{Relative Value Prediction.}
Given the resulting features for a pair of videos, $P_1$ and $P_2$,
we concatenate them and formulate a binary classification problem, indicating which video has a larger physical property value via a MLP network $\gamma$:
\begin{equation}
    \xi = \gamma ([P_1, P_2])
\end{equation}
The binary prediction for all three tasks is trained with binary cross entropy loss with the binary ground truth.

\subsection{Multimodal Large Language Models}
\label{sec:arch_language}

This section studies off-the-shelf multimodal large language models (MLLMs) for understanding dynamic physical properties from video. 
We explore various prompting strategies on state-of-the-art MLLMs, including Qwen2.5-VL-Max~\citep{hui2024qwen2}, GPT-4o~\citep{hurst2024gpt}, and Gemini 2.5 Pro~\citep{comanici2025gemini}, as illustrated in Figure~\ref{fig:arch} (right). 
Examples of the prompting strategies are provided in Appendix Section~\ref{sec:example_mllm}.

\subsubsection{Preliminary} 
The MLLM receives video frames as visual input. 
The text prompt includes (1) a brief description of the target property—for example: ``we are studying the viscosity of the liquid, where water is 1.0 and honey is 5000.0.'' 
This is followed by (2) a query, such as: ``what is the viscosity value of the liquid in the video?''~(absolute) or ``which video shows a liquid with higher viscosity? please output a decision score between 0 and 1, indicating the likelihood that the first video exhibits a higher property value.''~(relative). 
All the following prompt strategies provide (1) and (2) by default.

\subsubsection{Baseline Prompt} 

For {\em relative} tasks, we specify that the first $n$ frames belong to the first video and the last $n$ to the second. 
For {\em absolute} tasks, the default prompt is used.
Appendix Figure~\ref{fig:mllm_example_abs_0} and Figure~\ref{fig:mllm_example_rel_0}  provide an example of \textit{baseline prompt} for the absolute formulation and the relative formulation, respectively.

\subsubsection{Oracle Estimation Teaching}

For both \emph{relative} and \emph{absolute} settings, we provide the key cue to concentrate on from the Section~\ref{sec:arch_oracle} description to teach the MLLM how to estimate the properties step by step.
Appendix Figure~\ref{fig:mllm_example_abs_1} and Figure~\ref{fig:mllm_example_rel_1} provide an example of \textit{oracle estimation teaching} for the absolute formulation and the relative formulation, respectively.

\subsubsection{Few-Shot Examples} 

For both {\em relative} and {\em absolute} settings, we provide several examples, including the video input and desired ground truth. For fair comparison with visual prompting, we use examples in the synthetic training split.
Appendix Figure~\ref{fig:mllm_example_abs_2} and Figure~\ref{fig:mllm_example_rel_2} provide an example of \textit{few-shot examples} for the absolute formulation and the relative formulation, respectively.

\subsubsection{Frame Index Provided}

For both \emph{relative} and \emph{absolute} settings, we input the text of the index of each frame along with the frames. In this way the MLLMs may have a better understanding about the temporal relations between the input video frames.
Appendix Figure~\ref{fig:mllm_example_abs_3} and Figure~\ref{fig:mllm_example_rel_3} provide an example of \textit{frame index provided} for the absolute formulation and the relative formulation, respectively.

\subsubsection{Black Frames in Between} 

This strategy is only used for the {\em relative} setting. We insert black frames between the two video segments to clearly separate them. In the prompt, we refer to the videos as the frames before and after the black frames, rather than as the first and last $n$ frames.
Appendix Figure~\ref{fig:mllm_example_rel_4} provides an example of \textit{black frames in between} for the relative formulation. 

%% file: sec/5_experiment.tex
\section{Experiments}
\label{sec:exp}

\begin{figure*}[t]
    \centering
\includegraphics[width= 0.99\textwidth]{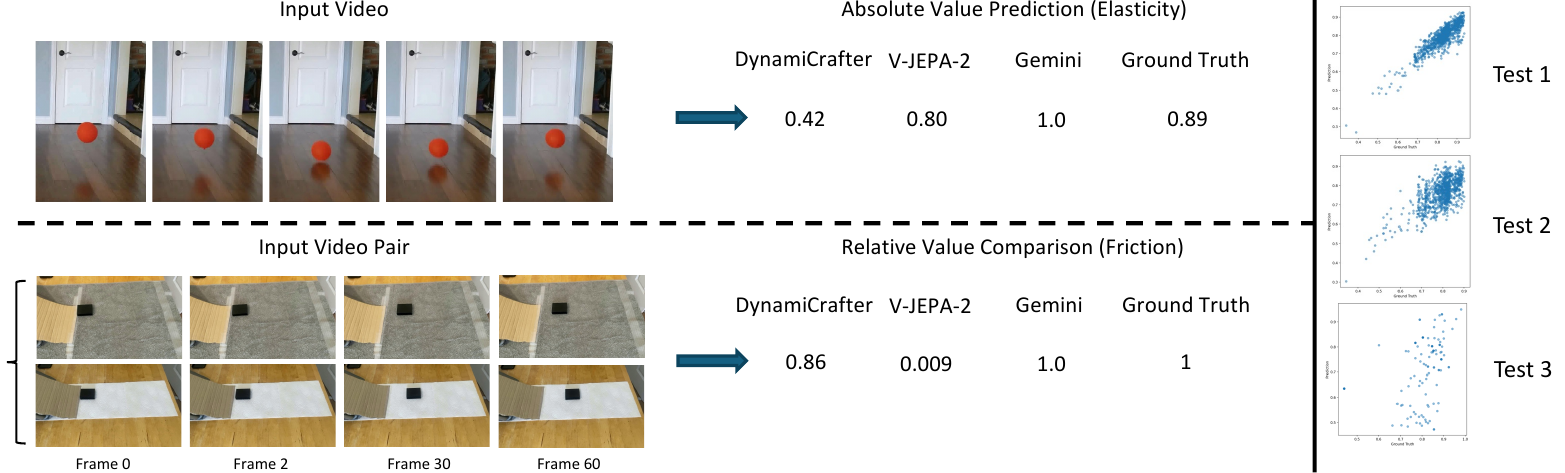}
% \vspace{-4pt}
\caption{
\textbf{Qualitative results.} 
\textbf{Top Left}: An example for elasticity absolute value prediction; 
\textbf{Bottom Left}: An example for friction relative value comparison. 
For each example, the original input video is shown on the left. 
Model predictions are shown on the right, including results from the Video Generative Model (DynamiCrafter), Video Self-Supervised Model (V-JEPA-2), and a MLLM (Gemini).
For the relative formulation, the ground truth value of `$1$' indicates that the first (top) video has larger dynamic friction coefficient than the second video.
In this example, the initial velocity of the lego brick in the two videos is similar (note the same displacement from frame $0$ to $2$), but the velocity reduces to $0$ at frame $30$ in the first video, while the object is still moving in frame $30$ to $60$ in the second video. 
\textbf{Right}: Scatter plots of prediction vs ground truth for the elasticity property from the V-JEPA-2 model. 
} 
% \vspace{-4mm}
\label{fig:prediction_comparison}
\end{figure*}

\noindent \textbf{Implementation Details.}
During oracle estimation, we train the GRU network with a learning rate of $1e^{-3}$ and the batch size is 128. For the generative and self-supervised video models, the backbones are frozen, 
the trainable parameters are optimised with a learning rate of $1e^{-5}$ and the batch size 16. 
For MLLMs, we perform  promp selection, and use the best strategy that we find for each of the absolute and relative settings for the experiments. 
\emph{Few-shot examples} and \emph{oracle estimation teaching} work best for the absolute and relative settings, respectively, as they directly provide the model with more context information about the properties.
Please refer to Appendix Section~\ref{sec:ablation_mllm} for the comparison results and analysis. All models are trained on H100/A6000/A40 GPUs. Please refer to Appendix Section~\ref{sec:additional_implementation_details} for more implementation details.

\noindent \textbf{Evaluation Metrics.} 
For {\em relative value comparison}, we report the ROC AUC score; 
for {\em absolute value prediction}, we use the Pearson Correlation Coefficient between the prediction and ground truth as this automatically calibrates the predictions to the scale of the ground truth. Please refer to Appendix Section~\ref{sec:supple_imple_detail_evaluation_metric} for more details and motivations on the evaluation metrics.

\subsection{Results for Relative Value Comparison}

Table~\ref{table:relative_absolute_combined} (left) shows relative value comparison results across physical properties and model types. 
The oracle estimator performs nearly perfectly on \texttt{test-1} and \texttt{test-2}, 
and strongly on \texttt{test-3}, indicating that the task is largely solvable using visual cues, geometry, and physics.
Both the evaluated generative (DynamiCrafter) and self-supervised (V-JEPA-2) video models achieve strong results on synthetic splits (\texttt{test-1} and \texttt{test-2}). Notably, they can also generalize well to the real-world split (\texttt{test-3}) for most scenarios of viscosity and elasticity, which rely on simple height ratios and expansion. However, friction proves more challenging for V-JEPA-2. Trained on synthetic data, it struggles to generalize, likely due to the fact that reliance on visual references ({\em e.g.}, ground plane grids) is absent in real videos, and due to friction’s inherent complexity involving higher-order motion and projective geometry of the viewpoint.
To further confirm, we introduce an additional real-world training split for friction videos with disjoint objects and surfaces from the test set (see Appendix Section~\ref{sec:supple_real_dataset_details} for more details). Fine-tuning the visual prompting architecture on this data improves performance on the real test split, as shown by the * values in Table~\ref{table:relative_absolute_combined}. 
Multimodal large language models (MLLMs), though not working very well with \emph{Baseline Prompt} (see Appendix Section~\ref{sec:ablation_mllm}), when prompted properly, also perform well, especially on real videos, which are more \emph{in-distribution} for them -- while on synthetic splits, their performance drops significantly.
This is likely due to the fact that the models tend to leverage semantic cues, \emph{e.g.,} the type of liquid or the category of object and surface, rather than visual motion.

\begin{table*}[h]
\setlength{\tabcolsep}{17pt}
\centering
\caption{
\textbf{Results for relative value comparison and absolute value prediction.} 
Left: ROC AUC scores for relative comparisons (range $[0, 1]$). 
Right: Pearson correlation coefficients for absolute predictions (range $[-1, 1]$). 
* indicates results after domain adaptation using a disjoint real training set.
\texttt{test-1} is the synthetic in-distribution test split; \texttt{test-2} is the synthetic out-of-distribution test split; \texttt{test-3} is the real-world test split.
}
% \vspace{-7pt}
\scalebox{0.8}{
\begin{tabular}{l|l|ccc||ccc}
\toprule
\multirow{2}{*}{\textbf{Property}} & \multirow{2}{*}{\textbf{Method}} 
& \multicolumn{3}{c||}{\textbf{Relative -- ROC AUC}} 
& \multicolumn{3}{c}{\textbf{Absolute -- Pearson Corr.}} \\
 &  & \textbf{Test-1} & \textbf{Test-2} & \textbf{Test-3} & \textbf{Test-1} & \textbf{Test-2} & \textbf{Test-3} \\
\midrule

\multirow{6}{*}{Elasticity} 
& \textbf{Oracle}              & 1.00 & 0.99 & 1.00 & 0.99 & 0.98 & 0.87  \\
\cmidrule{2-8}
& DynamiCrafter       & 1.00 & 0.99 & 0.61 & 0.92 & 0.78 & 0.10  \\
& V-JEPA-2 & 1.00 & 1.00 & 0.99 & 0.89 & 0.65 & 0.37 \\
\cmidrule{2-8}
& Qwen2.5VL-max                & 0.59 & 0.50 & 0.54 & -0.05 & 0.11 & 0.16 \\
& GPT-4o                       & 0.51 & 0.66 & 0.62  & 0.19 & 0.11 & 0.30  \\
& Gemini-2.5-pro              & 0.64 & 0.80 & 0.47 & 0.04 & 0.15 & 0.24 \\
\midrule

\multirow{6}{*}{Viscosity} 
& \textbf{Oracle}              & 0.99 & 0.99 & 1.00 & 0.99 & 0.98 & 0.80 \\
\cmidrule{2-8}
& DynamiCrafter       & 1.00 & 1.00 & 0.93  & 0.99 & 0.95 & 0.44  \\
& V-JEPA-2 & 1.00 & 0.94 & 0.99  & 0.98 & 0.69 & 0.61  \\
\cmidrule{2-8}
& Qwen2.5VL-max                & 0.64 & 0.61 & 0.86  & 0.16 & 0.06 & 0.02  \\      
& GPT-4o                       & 0.63 & 0.59 & 0.99  & 0.18 & 0.08 & 0.55  \\
& Gemini-2.5-pro              & 0.48 & 0.69 & 0.95  & -0.06 & -0.05 & 0.60  \\
\midrule

\multirow{8}{*}{Friction} 
& \textbf{Oracle}              & 1.00 & 1.00 & 0.87  & 0.99 & 1.00 & 0.83  \\
\cmidrule{2-8}
& DynamiCrafter       & 0.96 & 0.90 & 0.97  & 0.95 & 0.78 & 0.25  \\
& \quad + Domain Adaptation    &  --  &  --  & --   &  --  &  --  & 0.80*  \\
& V-JEPA-2 & 0.97 & 0.90 & 0.48  & 0.87 & 0.56 & 0.21  \\
& \quad + Domain Adaptation    &  --  &  --  & 0.87*   &  --  &  --  & 0.73*  \\
\cmidrule{2-8}
& Qwen2.5VL-max                & 0.52 & 0.48 & 0.80  & -0.03 & 0.03 & 0.06  \\
& GPT-4o                       & 0.49 & 0.40 & 0.67  & 0.07 & 0.10 & 0.38  \\
& Gemini-2.5-pro               & 0.56 & 0.52 & 0.97  & 0.01 & 0.10 & 0.12  \\
\bottomrule
\end{tabular}
}
% \vspace{-4mm}
\label{table:relative_absolute_combined}
\end{table*}

\subsection{Results for Absolute Value Prediction}

Table~\ref{table:relative_absolute_combined}~(right) shows results for absolute value prediction across physical properties and methods. This task is more challenging than relative comparison, as models must regress quantitative physical values rather than compare video pairs from the same viewpoint. Similar to the relative setting, the oracle estimator achieves near-perfect performance on \texttt{test-1} and \texttt{test-2}, and strong performance on \texttt{test-3}, confirming that the task is largely solvable through visual cues, multi-view geometry, and physical laws.
We highlight several key observations:
(i) \textbf{comparable performance across backbones}.
Despite being trained for generative tasks, DynamiCrafter performs on par with V-JEPA-2 when predicting dynamic physical properties.
(ii) \textbf{friction remains challenging}. 
Similar to the relative setting, both DynamiCrafter and V-JEPA-2 struggle with friction estimation. Performance again improves with domain adaptation.
(iii) \textbf{MLLMs better on real test split than synthetic}. 
MLLMs continue to perform better on the real test split than synthetic test splits, benefiting from their familiarity with real-world visual semantics.
(iv) \textbf{greater gap from oracle}. 
The performance gap between video foundation models and the oracle is more pronounced here than in the relative setting, indicating that accurate physical value regression remains a significant challenge for current video models.

\subsection{Qualitative Results}
\label{sec:exp_qualitative}

Figure~\ref{fig:prediction_comparison} (left) shows qualitative examples comparing model predictions across different tasks.
In the \textbf{first row}, we illustrate an example from the elasticity absolute value prediction task. The video generative model (DynamiCrafter), self-supervised model (V-JEPA-2), and MLLM (Gemini) predict values of 0.42, 0.80, and 1.0, respectively—all reasonably close to the ground-truth value of 0.89.
In the \textbf{second row}, we present a friction relative value comparison task. The input consists of two videos, where the first exhibits a higher dynamic friction coefficient than the second. 
Both the video generative model and the MLLM correctly assign high likelihoods to this relationship (0.86 and 1.0, respectively), aligning with the ground truth. In contrast, the self-supervised model incorrectly predicts the reverse and does so with high confidence.
Figure~\ref{fig:prediction_comparison} (right) shows examples of the scatter plots for the absolute value prediction. More specifically, we show the scatter plots of video self-supervised model on the three test splits. It can be observed that the performance degrades from \texttt{test-1} to \texttt{test-3}, as \texttt{test-1} is of the same distribution as the synthetic training split, while \texttt{test-2} is out-of-distribution synthetic test and \texttt{test-3} is for real evaluation. We provide more scatter plots in Appendix Section~\ref{sec:additional_scatter_plot}.

%% file: sec/6_conclusion.tex
\section{Conclusion}
\label{sec:conclusion}

We investigate the task of inferring dynamic physical properties—elasticity, viscosity, and friction—from videos. To support this, we introduce a benchmark dataset with ground-truth annotations and evaluate representative video foundation models under both absolute prediction and relative comparison settings. 
We adopt a simple architecture to extract physical cues from off-the-shelf generative and self-supervised video models, and explore prompting strategies to elicit predictions from MLLMs.
Experiments show the evaluated generative and self-supervised models have generally similar and reasonable performance, though this evaluation should be extended to further models in the future. 
MLLMs perform worse overall but improve with more informative prompting, 
especially on real-world data. 
The worse performance of MLLMs is consistent with   previous work~\cite{fu2025hidden}, where it is observed that visual information is not properly fused in the language model.
However, all evaluated models fall short of the oracle, particularly in absolute value prediction. 
These results highlight the need to enhance physical reasoning in video models—a key direction for future research.

%% file: sec/7_supple.tex
\onecolumn

{
    \centering
    \Large
    \textbf{Appendix}\\
    \vspace{1.0em}
}

\section{Additional Implementation Details}
\label{sec:additional_implementation_details}
\vspace{5pt}

As mentioned in Section~\ref{sec:exp} of the main paper, in this section we provide more implementation details. Please refer to our code in the .zip file for more details.

\subsection{Computing Infrastructure}
\label{sec:supple_imple_detail_computing_infra}
\vspace{3pt}

The video generative and video self-supervised models are trained on a single H100/A6000/A40 GPU. The models are trained much faster on H100, and much slower on A6000 and A40, but all of them can be used to train and inference our models.
The GPU memory of H100 GPU is 96GB, and 48GB for A6000/A40 GPU. 
The CPU memory associated with the GPU is 120GB for the H100 GPU, and 90GB for A6000/A40 GPU. 
The experiments are conducted on a Linux cluster. We provide the names and versions of the relevant software libraries and frameworks in the code.

\subsection{Dimension of Video Feature Extraction Vectors and Details of MLPs}
\label{sec:supple_imple_detail_dimension_mlp}
\vspace{3pt}

We set the dimension of the learnable query vector $q$ to be $2560$ for DynamiCrafter or V-JEPA-2. 
For DynamiCrafter, $q$ is mapped to the dimensions of different layers for cross-attention with different layers of pre-trained feature tokens, by a set of MLPs 
$
f_1^{G} : \mathbb{R}^{2560} \;\rightarrow\;
\mathbb{R}^{(320^{\times 4},\,640^{\times 3},\,1280^{\times 12},\,640^{\times 3},\,320^{\times 3})}
$; 
For V-JEPA-2, $q$ is mapped to the dimensions of different layers for cross-attention with different layers of pre-trained feature tokens, by a set of MLPs 
$
f_1^S : \mathbb{R}^{2560} \;\rightarrow\; \mathbb{R}^{(1024^{\times 16})}
$. 
The resulting vectors $p$ therefore can be represented as $\mathbb{R}^{(320^{\times 4},\,640^{\times 3},\,1280^{\times 12},\,640^{\times 3},\,320^{\times 3})}$ for DynamiCrafter and $\mathbb{R}^{(1024^{\times 16})}$ for V-JEPA-2. The $p$ vectors are then mapped by another set of MLPs
$
f_2^G : \mathbb{R}^{(320^{\times 4},\,640^{\times 3},\,1280^{\times 12},\,640^{\times 3},\,320^{\times 3})}
\;\rightarrow\; \mathbb{R}^{(2560^{\times 25})}
$
for DynamiCrafter, and
$
f_2^S : \mathbb{R}^{(1024^{\times 16})} \;\rightarrow\; \mathbb{R}^{(2560^{\times 16})}
$
for V-JEPA-2 and then average pooled to get $P \in \mathbb{R}^{2560}$.

\subsection{Details of Video Input}
\label{sec:supple_imple_detail_video_input}
\vspace{3pt}

We uniformly sample 16 frames per video as input to all the models for fair comparison. The 16 frames are uniformly sampled so that the physics process we want to study is properly reflected with the sampled 16 frames, \emph{e.g.,} the dropping and bouncing of the ball, the expansion of the liquid, and the slowing-down sliding process of the object. For construction of relative pairs of videos, we randomly get a list of different viewpoints first, and then for each viewpoint, we randomly generate $m$ videos and then
randomly sample pairs from $m \times (m-1)$ possible video pairs. The binary ground truth for the pair is obtained via comparing the property values of the two videos.

\subsection{Motivation and Details of Evaluation Metrics}
\label{sec:supple_imple_detail_evaluation_metric}
\vspace{3pt}

We use the ROC AUC score for the relative formulation, as it is a binary classification problem, and AUC is a good evaluation metric to reflect the model's performance over different decision thresholds. ROC AUC is computed as the area under the receiver-operating-characteristic curve, \emph{i.e.}, the integral of the true-positive rate versus the false-positive rate over all possible classification thresholds. We use the Pearson Correlation Coefficient for the absolute formulation, as it can reflect how the model's predicted values correlate to the ground truth values -- and our goal here is the \emph{value ordering}, rather than their absolute prediction, as accurately determining the ground truth values is difficult, particularly for viscosity. 
Also, the correlation is more forgiving than absolute error for out of distribution prediction (between the train and test sets).

\subsection{Segmentation Masks and Corner Detections for Oracle Estimation}
\label{sec:supple_imple_detail_oracle_seg_corner}
\vspace{3pt}

In the oracle estimation, we need to segment the target object or liquid. For the synthetic videos, the segmentation masks can be directly obtained from the simulator. For the real videos, we obtain the segmentations by using the pre-trained Grounded 
SAM~2~\citep{ravi2024sam2segmentimages,liu2023grounding,ren2024grounding,ren2024grounded,kirillov2023segany}, with text prompts such as `falling ball' (for elasticity), and `sliding' + the names of different sliding objects (for friction). 
For viscosity,  we only need the mask of the liquid on the plate surface, but the presence of the liquid column interferes -- and directly using the liquid’s name as a prompt for Grounded SAM 2 makes segmentation difficult. Therefore, we first segment the plate, and then apply morphological processing (taking the enclosed central region of the plate and using a closing operation to remove the liquid columns) to obtain the mask of the liquid on the plate surface.
After getting the automatically predicted masks, we manually filter and adjust them to make sure they are of good quality. 
Apart from segmentation masks, for the friction oracle estimation, we also need to detect the corner points of the sliding cube. For the synthetic videos, we annotate the corners with a different color so we can easily detect them. For the real videos, we annotated the corner positions manually as they are  difficult to obtain automatically.

\clearpage
\section{Dataset Details}
\label{sec:supple_dataset_details}
\vspace{5pt}

As mentioned in Section~\ref{sec:problem} of the main paper, in this section we provide further details regarding the collection of synthetic and real datasets.

\subsection{Synthetic Datasets}

As described in Section~\ref{sec:problem} of the main paper, the simulator uses two distinct domains of nuisance parameters: $\mathcal{A}_1$ and $\mathcal{A}_2$. The \texttt{train} and \texttt{test-1} splits are generated by sampling from $\mathcal{A}_1$, while \texttt{test-2} is generated from $\mathcal{A}_2$. Below, we detail the differences between these domains for each dynamic physical property.

The Genesis simulator operates in a world coordinate system where gravity points in the $-z$ direction, and physical processes are centered around the origin $(0,0,0)$. The camera position is defined by three parameters: height $h$ (controls the $z$ coordinate), radius $R$ (distance from the $(0,0)$ point in the $xy$-plane), and angle $\alpha$ (deviation from the $+x$ direction). Camera orientation is further specified by the 3D point $(x_l, y_l, z_l)$ that the camera looks at.
Object and liquid colors are defined by RGB values $(r, g, b)$. Lighting remains fixed from the $+x$ direction, meaning changes in camera viewpoint also affect lighting conditions on the object.

Table~\ref{tab:param_range_elasticity}, Table~\ref{tab:param_range_viscosity} and Table~\ref{tab:param_range_friction} detail the parameter settings for each physical property. All parameter ranges are chosen to ensure the visibility of the studied phenomena—e.g., the drop-and-bounce motion of a ball—in the synthetic videos.

\begin{table}[h]
\centering
\caption{\textbf{Parameter Ranges for Elasticity}. Values are randomly sampled per domain if it is a range. Top: Nuisance parameters; Bottom: The target dynamic physical property we study.}
\begin{tabular}{l|c|c}
\toprule
\textbf{Parameter} & \textbf{$\mathcal{A}_1$} & \textbf{$\mathcal{A}_2$} \\
\midrule
$R$ & $1.5$ & $1.5$ \\
$h$ & $(0.5, 1.5)$ & $(0.25, 0.5)$ \\
$\alpha$ & $(0, \frac{1}{2}\pi)$ & $(\frac{1}{2}\pi, 2\pi)$ \\
$x_l$ & $(-0.1, 0.1)$ & $(0.1, 0.2)$ \\
$y_l$ & $(-0.1, 0.1)$ & $(-0.2, -0.1)$ \\
$z_l$ & $(0.05, 0.27)$ & $(-0.05, 0.05)$ \\
$r$ & $(0, 1)$ & $0$ \\
$g$ & $(0, 1)$ & $0$ \\
$b$ & $0$ & $(0, 1)$ \\
Drop height & $(0.25, 0.4)$ & $(0.4, 0.5)$ \\
Ball radius & \multicolumn{2}{c}{$0.1$} \\
\midrule
Elasticity & \multicolumn{2}{c}{$(0, 1)$} \\
\bottomrule
\end{tabular}

\label{tab:param_range_elasticity}
\end{table}

\begin{table}[h]
\centering
\caption{\textbf{Parameter Ranges for Viscosity}. Values are randomly sampled per domain if it is a range. Top: Nuisance parameters; Bottom: The target dynamic physical property we study.}
\begin{tabular}{l|c|c}
\toprule
\textbf{Parameter} & \textbf{$\mathcal{A}_1$} & \textbf{$\mathcal{A}_2$} \\
\midrule
$R$ & $1.5$ & $1.5$ \\
$h$ & $(0.5, 1.5)$ & $(0.25, 0.5)$ \\
$\alpha$ & $(0, \frac{1}{2}\pi)$ & $(\frac{1}{2}\pi, 2\pi)$ \\
$x_l$ & $(-0.1, 0.1)$ & $(0.1, 0.2)$ \\
$y_l$ & $(-0.1, 0.1)$ & $(-0.2, -0.1)$ \\
$z_l$ & $(0.05, 0.27)$ & $(-0.05, 0.05)$ \\
$r$ & $(0, 1)$ & $0$ \\
$g$ & $(0, 1)$ & $0$ \\
$b$ & $0$ & $(0, 1)$ \\
Drop height & \multicolumn{2}{c}{$0.056$} \\
Liquid column height & \multicolumn{2}{c}{$0.1$} \\
Liquid column radius & \multicolumn{2}{c}{$0.05$} \\
\midrule
Viscosity & \multicolumn{2}{c}{$(5\mathrm{e}{-5}, 1\mathrm{e}{-2})$} \\
\bottomrule
\end{tabular}

\label{tab:param_range_viscosity}
\end{table}

\begin{table}[h]
\centering
\caption{\textbf{Parameter Ranges for Friction}. Values are randomly sampled per domain if it is a range. Top: Nuisance parameters; Bottom: The target dynamic physical property we study. $(x_0, y_0)$: initial position of the sliding cube; $(v^x_{0}, v^y_{0})$: initial velocity of the sliding cube.}
\begin{tabular}{l|c|c}
\toprule
\textbf{Parameter} & \textbf{$\mathcal{A}_1$} & \textbf{$\mathcal{A}_2$} \\
\midrule
$R$ & $1.5$ & $1.5$ \\
$h$ & $(0.5, 1.5)$ & $(0.25, 0.5)$ \\
$\alpha$ & $(0, \frac{1}{2}\pi)$ & $(\frac{1}{2}\pi, 2\pi)$ \\
$x_l$ & $(-0.1, 0.1)$ & $(0.1, 0.2)$ \\
$y_l$ & $(-0.1, 0.1)$ & $(-0.2, -0.1)$ \\
$z_l$ & $(-0.1, 0.12)$ & $(-0.14, -0.1)$ \\
$r$ & $(0, 1)$ & $(0, 1)$ \\
$g$ & $(0, 1)$ & $(0, 1)$ \\
$b$ & $0$ & $(0, 1)$ \\
$x_0$ & $(-0.1, 0.1)$ & $(-0.15, -0.1)$ \\
$y_0$ & $(-0.1, 0.1)$ & $(0.1, 0.15)$ \\
$v_0^x$ & $(0.6, 1.0)$ & $(1.0, 1.2)$ \\
$v_0^y$ & $(0.6, 1.0)$ & $(1.0, 1.2)$ \\
Motion direction & \multicolumn{2}{c}{$v_0^x$ or $v_0^y$ with prob. $0.5$} \\
Cube size & \multicolumn{2}{c}{$0.1$} \\
\midrule
Friction coeff. & \multicolumn{2}{c}{$(0, 0.2)$} \\
\bottomrule
\end{tabular}

\label{tab:param_range_friction}
\end{table}

\subsection{Real Datasets Details}
\label{sec:supple_real_dataset_details}

\vspace{3pt} \noindent \textbf{Elasticity Dataset.} 
This dataset contains video clips sourced from the Internet, capturing a variety of ball types being dropped—{\em e.g.}, basketball, tennis ball, soccer ball, rubber ball, balloon (air-filled), exercise ball, medicine ball, marble, and tomato. Ground truth elasticity values range from 0.44 (tomato) to 0.98 (tennis ball).

\vspace{3pt} \noindent \textbf{Viscosity Dataset.} 
We include 12 different liquids: coffee, vinegar, cola, wine, cooking wine, whole milk, hot chocolate, dark soy sauce, smoothie, sesame oil, cream, and maple syrup. Ground truth viscosity values are obtained from online sources. For cases where the value is reported as a range, we use the midpoint as the ground truth. Values range from 1.2 (coffee) to 225 (maple syrup).

\vspace{3pt} \noindent \textbf{Friction Dataset.} 
This dataset contains 5 objects—plastic disk, plastic LEGO brick, paper box, metal pencil box, and wooden box—and 6 surfaces: gray towel, kitchen paper, tablecloth, red towel, wooden table, and cardboard.

The training split includes:
\begin{itemize}
    \item \textbf{Objects:} paper box, metal pencil box, wooden box
    \item \textbf{Surfaces:} red towel, wooden table, cardboard
\end{itemize}

The testing split includes:
\begin{itemize}
    \item \textbf{Objects:} plastic disk, LEGO brick
    \item \textbf{Surfaces:} gray towel, kitchen paper, tablecloth
\end{itemize}

\begin{figure*}[h]
    \centering
\includegraphics[width=\textwidth]{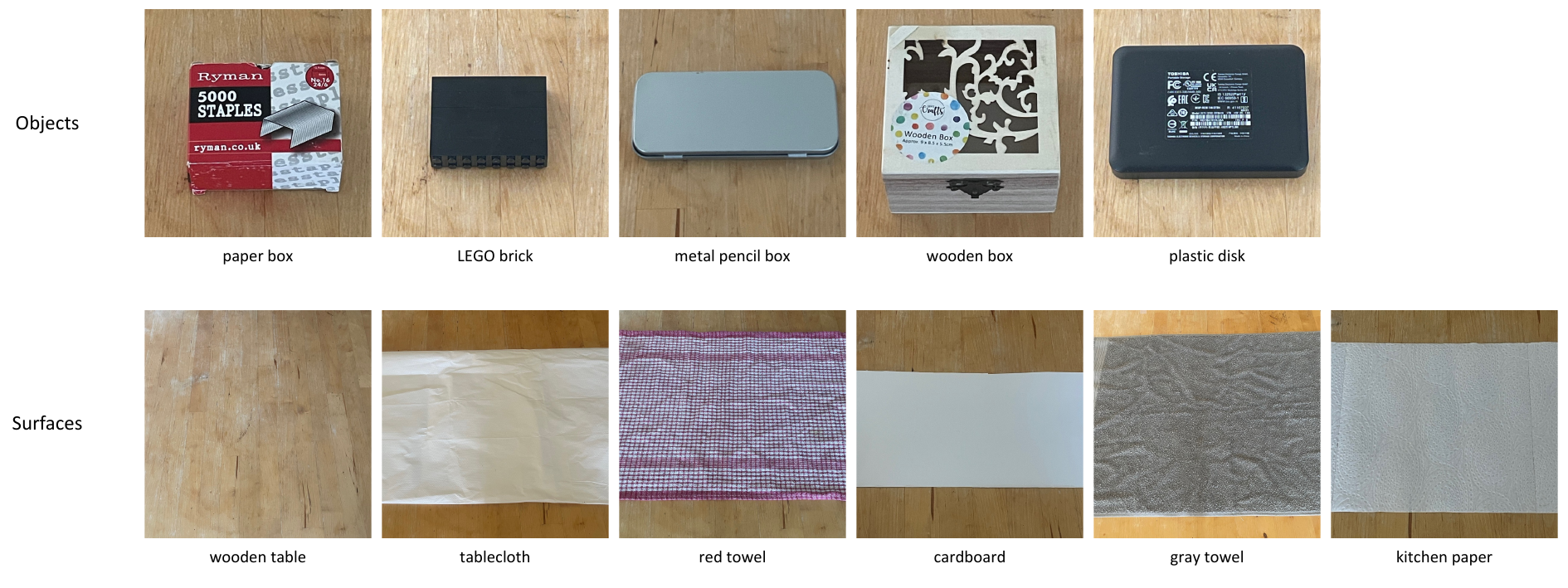}
\caption{
\textbf{Objects and surfaces in the \emph{friction} real dataset.} 
Top: Objects used for friction real dataset collection; Bottom: Surfaces used for friction real dataset collection.
} 
\label{fig:object_and_surface}
\end{figure*}

Figure~\ref{fig:object_and_surface} shows close-up images of all objects and surfaces. Ground truth friction values range from 0.105 (pencil box on plastic paperboard) to 0.544 (LEGO on gray towel).
The ground truth dynamic friction coefficient values are measured using a spring dynamometer by dragging the object at constant speed, as mentioned in the main paper. 
All datasets will be made publicly available upon paper acceptance under the CC-BY-4.0 license.

\subsection{Devices for Real Dataset Collection}
\vspace{3pt}

\begin{figure*}[h]
    \centering
\includegraphics[width=\textwidth]{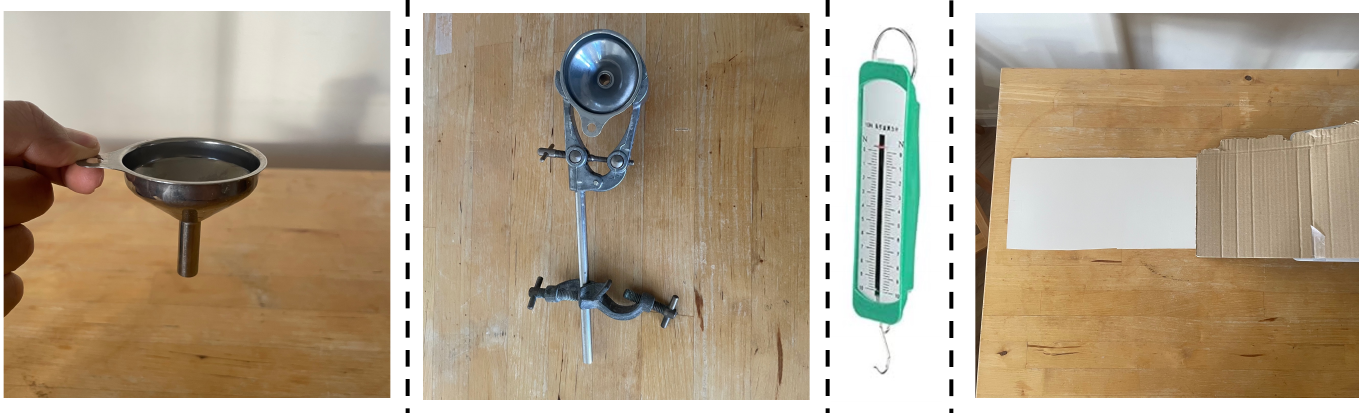}
\caption{
\textbf{Devices used to collect real datasets.} 
\textbf{Left}: The funnel used in the collection of the viscosity real dataset; 
\textbf{Middle-left}: The funnel holder used in the collection of the viscosity real dataset; 
\textbf{Middle-right}: The spring dynamometer used to measure the ground truth dynamic friction coefficient in the collection of the friction real dataset;
\textbf{Right}: The slope used to give the objects an initial velocity on the horizontal surface in the collection of the friction real dataset.
} 
\label{fig:device}
\end{figure*}

As mentioned in Section~\ref{sec:data} of the main paper, we show the devices that we used to collect our real datasets in Figure~\ref{fig:device}. The funnel (left) and the funnel holder (middle-left) are used to collect the viscosity real dataset. The spring dynamometer (middle-right) is used to measure the ground truth dynamic friction coefficient in the collection of the friction real dataset. The slope (right) is used in the friction experiment, where we slides the object down the slope to give it an initial velocity on the horizontal surface.

\clearpage

\section{Derivation of Oracle Estimations Equations}
\label{sec:derivation_physics}

\vspace{5pt}

As mentioned in Section~\ref{sec:data} of the main paper, in this section, we give the derivation for the oracle estimations in the main paper, based on standard definitions for the properties and the laws of physics. 
Note, we only require the oracle values for each property to be determined up to a common scale, since the correlation used to evaluate performance is unaffected by the overall scale.

\vspace{10pt} \noindent \textbf{Elasticity.} 
The {\em coefficient of elasiticity} or {\em coefficient of restitution} $e$ is defined as
\begin{equation*}
    e = \frac{v_{\text{before impact}}}{v_{\text{after impact}}} 
\end{equation*}
where $v$ is the magnitude of the velocity. In our case the ball is dropped from rest at a height 
$h_{\text{drop}}$ onto a horizontal surface, and bounces in the vertical direction to the height $h_{\text{bounce}}$. It can be shown (see~\cite{wiki:cor}) that in this case:
\begin{equation}
    e = 
    \sqrt{\frac{h_{\text{bounce}}}{h_{\text{drop}}}}
\end{equation}

\vspace{10pt} \noindent \textbf{Viscosity.}
\emph{Viscosity} is a physical property that characterizes a fluid’s internal resistance to flow~\citep{wiki:vis}. In our case, we study the case that the liquid is expanding on the ground plane. According to the spreading dynamics~\citep{wiki:wet} of liquid, the radius (thus the area) of the liquid is an inverse function of the viscosity, given other parameters controlled, such as the density of the liquid $\rho$, the diameter (thus the volume) of the liquid column $D$, and the dropping velocity $v$ of the liquid column when it reaches the ground. In our case, we control $D$ as we use a funnel with a fixed nozzle diameter to produce a consistent liquid column and we always pour the same volume of liquid into the funnel; we control $v$ as we use a funnel holder that allows us to fix the height from which the liquid is poured; $\rho$ is roughly controlled as the liquids are of similar density, ranging from $0.92$ (\emph{e.g.,} sesame oil ) to $1.05$ (\emph{e.g.,} smoothie), except for maple syrup which is $1.32$, but as the ground truth viscosity of maple syrup is much higher than other liquids, this variation will not influence much when we calculate the Pearson Correlation Coefficient between predictions and ground truth values. 
Therefore, we assume 
\begin{equation}
    \mu \propto \frac{1}{(d(A(t)) / dt)^\alpha}
\end{equation}
where $\mu$ is the viscosity of the liquid, and $A(t)$ is the liquid area size as a function of time $t$.
In practice, we try with $\alpha=1$, \emph{i.e.,} $\mu \propto \frac{1}{d(A(t)) / dt}$ and gets reasonable oracle test results, so we set $\alpha=1$ for our oracle estimations.

\vspace{10pt} \noindent \textbf{Friction.} 
If $F$ is the dynamic friction force acting on the object, then the dynamic friction coefficient $\mu_k$ is defined by the equation $F = \mu_k \times $ normal force on the object. In our case, the object moves on a horizontal surface, and the normal force is the weight of the object, so $F = \mu_k m g $, where $m$ is the mass of the object, and $g$ is the gravity acceleration. From Newton's Second Law $F = m a $, we therefore have $ a = \mu_k g $, \emph{i.e.,} 
\begin{equation}
    \mu_k = \frac{a}{g}
\end{equation}
where $a$ is the acceleration of the object.

\clearpage
\section{Ablation for Different Strategies of MLLM prompting}
\label{sec:ablation_mllm}

\vspace{5pt}

As mentioned in Section~\ref{sec:exp} of the main paper, we conduct an ablation study on the elasticity task to identify the most effective prompting strategy for MLLMs, using \textbf{Gemini 2.5 Pro} due to its strong performance in video understanding and visual reasoning. Results for the absolute and relative formulations are shown in Table~\ref{tab:mllm_ablation_absolute} and Table~\ref{tab:mllm_ablation_relative}, respectively.
Due to the high computational cost of MLLM inference, we perform the ablation on a randomly selected subset of 20 samples per test split. The results show that the \textbf{Few-Shot Examples} strategy performs best for the absolute formulation, while \textbf{Oracle Estimation Teaching} is most effective for the relative formulation.

In Section~\ref{sec:example_mllm}, we provide examples for each of the prompting strategies and detailed analysis regarding the influence of each strategy to the final performance.

\begin{table}[!thb]
\centering

\caption{\textbf{Absolute prediction results for different MLLM prompting strategies}. We conduct the study on Gemini 2.5 Pro for the elasticity task.
}

\setlength{\tabcolsep}{15pt}

\begin{tabular}{l|cccc}
\toprule
Strategy & Test-1 & Test-2 & Test-3 & Avg \\
\midrule
Baseline                    & -0.03 & 0.26 & 0.06 & 0.10 \\
+ Frame Index Provided      & 0.06 & 0.35 & 0.55 & 0.32 \\
+ Few-Shot Examples         & 0.39 & 0.34 & 0.24 & \textbf{0.33} \\
+ Oracle Estimation Teaching & 0.19 & 0.14 & 0.24 & 0.19 \\
\bottomrule
\end{tabular}

\label{tab:mllm_ablation_absolute}
\end{table}

\vspace{0.5em}

\begin{table}[!thb]
\centering

\caption{\textbf{Relative comparison results for different MLLM prompting strategies}. We conduct the study on Gemini 2.5 Pro for the elasticity task.}

\setlength{\tabcolsep}{15pt}

\begin{tabular}{l|cccc}
\toprule
Strategy & Test-1 & Test-2 & Test-3 & Avg \\
\midrule
Baseline                    & 0.51 & 0.74 & 0.55 & 0.60 \\
+ Black Frames in Between   & 0.56 & 0.72 & 0.63 & 0.64 \\
+ Frame Index Provided      & 0.54 & 0.80 & 0.52 & 0.62 \\
+ Few-Shot Examples         & 0.43 & 0.65 & 0.52 & 0.54 \\
+ Oracle Estimation Teaching & 0.63 & 0.79 & 0.54 & \textbf{0.65} \\
\bottomrule
\end{tabular}

\label{tab:mllm_ablation_relative}
\end{table}

\clearpage
\section{Examples of Different Prompting Strategies}
\label{sec:example_mllm}

\vspace{5pt}
As mentioned in Section~\ref{sec:arch_language} of the main paper, in this section we provide examples of different prompting strategies for both the \textbf{absolute formulation} and the \textbf{relative formulation}.

\begin{figure*}[h]
    \centering
\includegraphics[width=\textwidth]{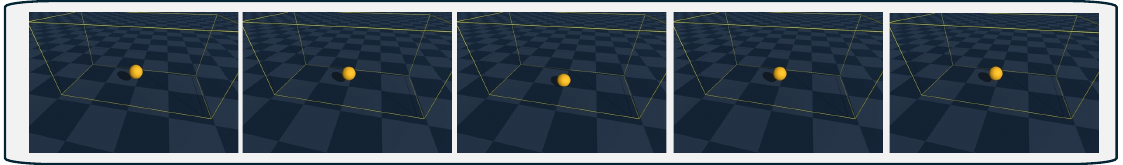}
\caption{
\textbf{Visual input for the \emph{absolute formulation}.} The example is on Gemini for the elasticity property. Five representative frames are shown for the input video.
} 
\label{fig:mllm_example_abs}
\end{figure*}

\begin{figure*}[h]
    \centering
\includegraphics[width=\textwidth]{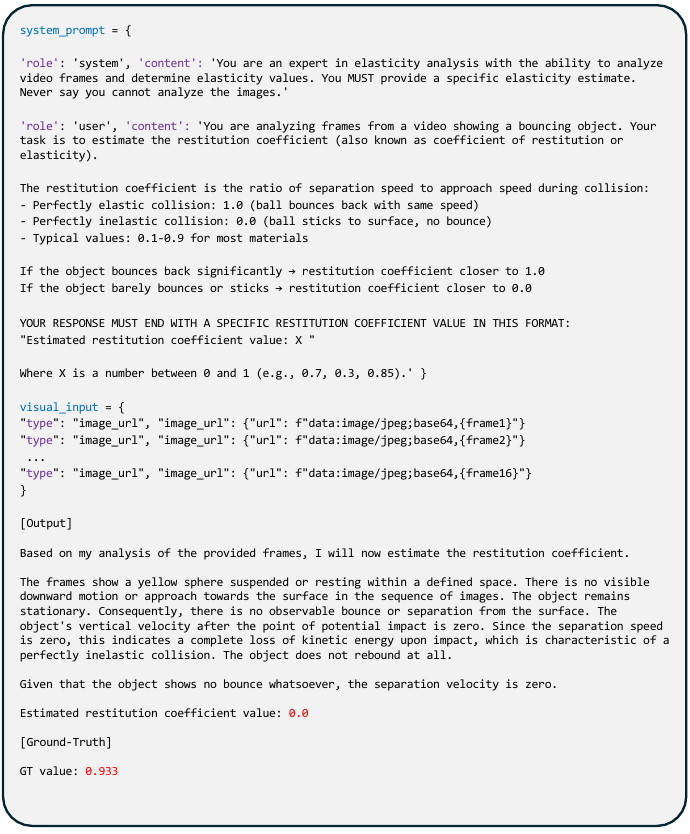}
\caption{
\textbf{Example of \emph{baseline prompt} for the \emph{absolute formulation}.} The example is on Gemini for the elasticity property. The initial state of object motion is incorrectly recognized from the beginning.
} 
\label{fig:mllm_example_abs_0}
\end{figure*}

\begin{figure*}[h]
    \centering
\includegraphics[width=0.7\textwidth]{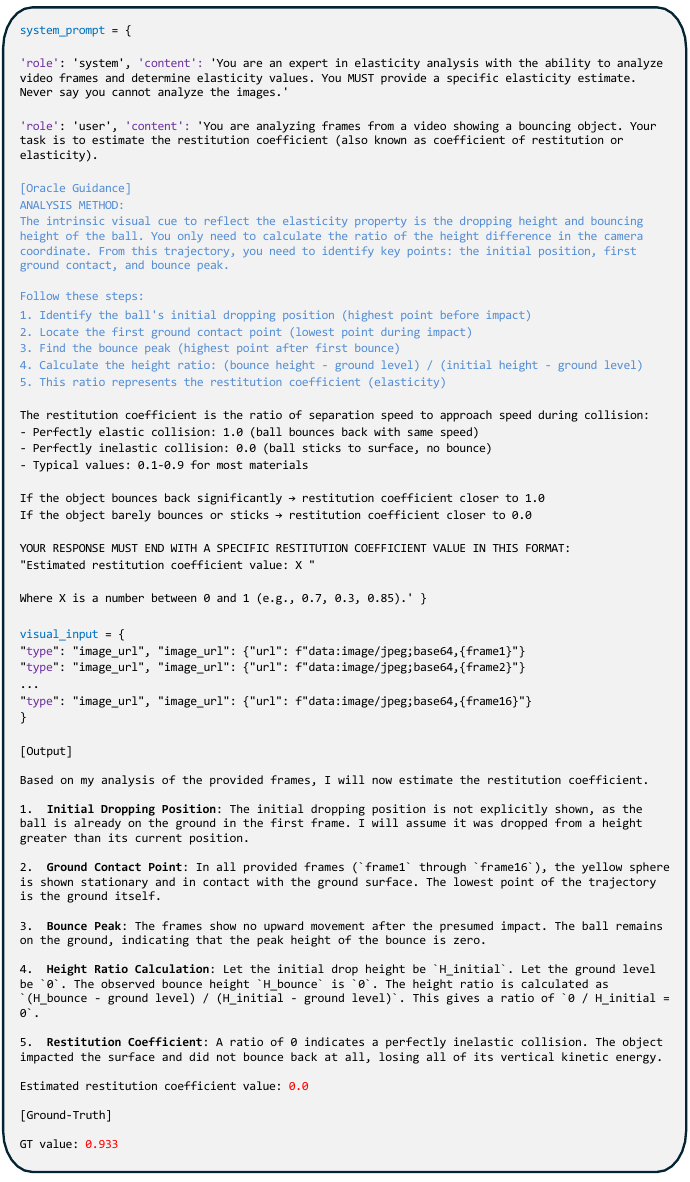}
\caption{
\textbf{Example of \emph{oracle estimation teaching} for the \emph{absolute formulation}.} The example is on Gemini for the elasticity property. Although the model strictly follows the oracle’s step-by-step guidance, an incorrect identification of the peak in the third step leads to a significantly inaccurate final prediction.
} 
\label{fig:mllm_example_abs_1}
\end{figure*}

\begin{figure*}[h]
    \centering
\includegraphics[width=0.7\textwidth]{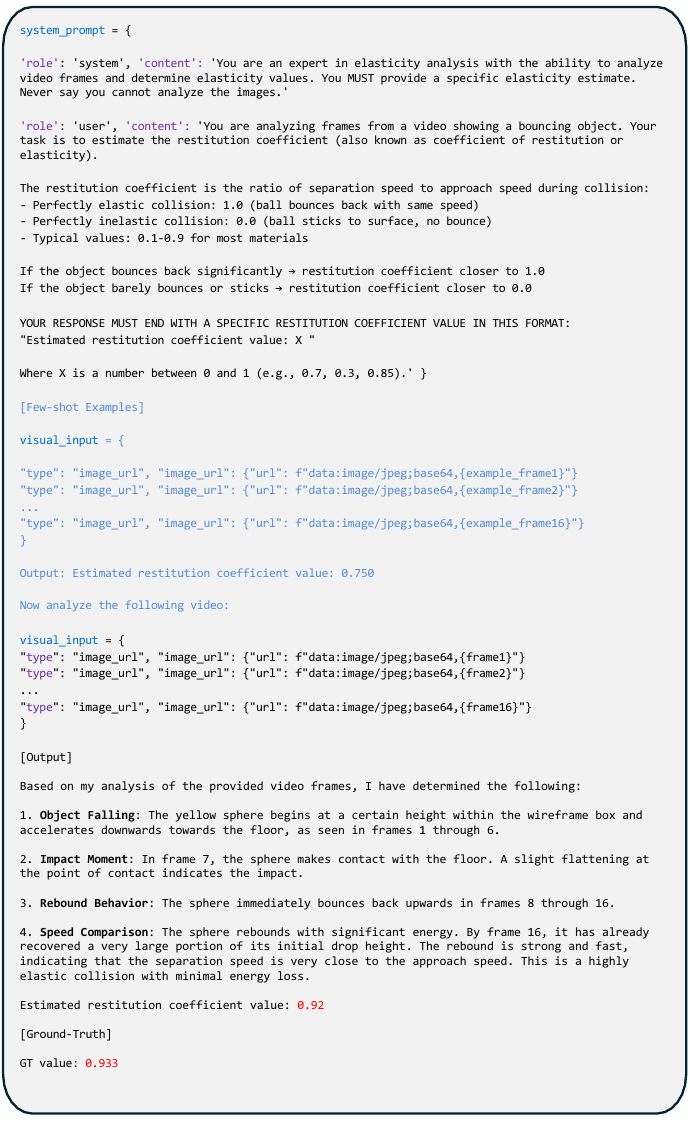}
\caption{
\textbf{Example of \emph{few-shot examples} for the \emph{absolute formulation}.} The example is on Gemini for the elasticity property. The ground truth examples provided in the few-shot setting serve as effective calibration signals, leading to notably improved performance.
} 
\label{fig:mllm_example_abs_2}
\end{figure*}

\begin{figure*}[h]
    \centering
\includegraphics[width=0.9\textwidth]{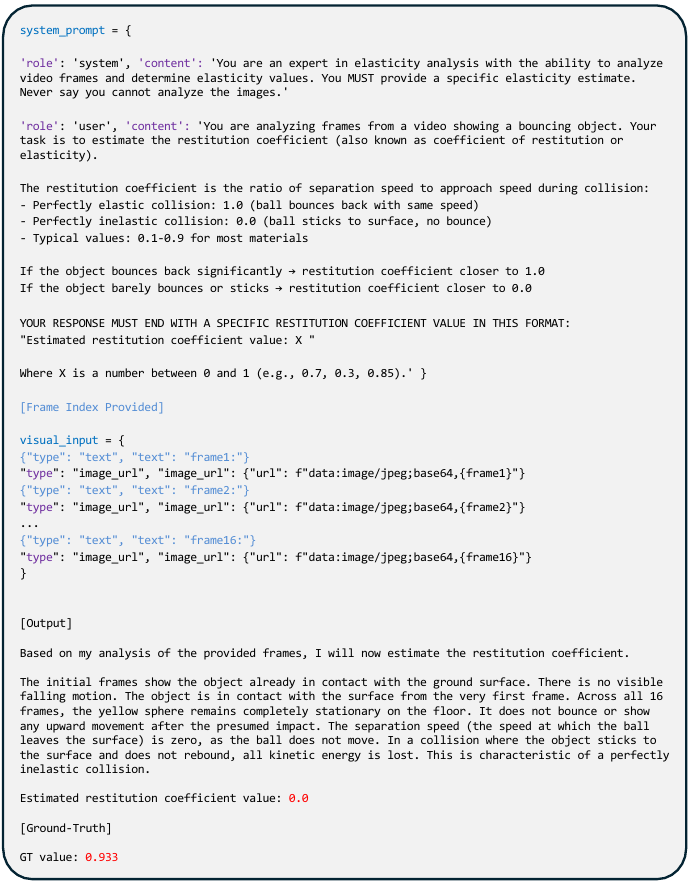}
\caption{
\textbf{Example of \emph{frame index provided} for the \emph{absolute formulation}.} The example is on Gemini for the elasticity property. Providing frame indices helps the model better interpret the motion process. However, estimating the final value based solely on this information remains challenging.
} 
\label{fig:mllm_example_abs_3}
\end{figure*}

The examples of \textbf{absolute formulation} are provided in Figure~\ref{fig:mllm_example_abs} to Figure~\ref{fig:mllm_example_abs_3}. More specifically, Figure~\ref{fig:mllm_example_abs} shows the visual input to the MLLM; Figure~\ref{fig:mllm_example_abs_0} shows the prompt and model output of \emph{baseline prompt}; 
Figure~\ref{fig:mllm_example_abs_1} shows the prompt and model output of \emph{oracle estimation teaching}; 
Figure~\ref{fig:mllm_example_abs_2} shows the prompt and model output of \emph{few-shot examples}; 
Figure~\ref{fig:mllm_example_abs_3} shows the prompt and model output of \emph{frame index provided}.

It can be observed that:
\begin{itemize}

\item \textbf{Baseline Prompt.}
The initial state of object motion is incorrectly recognized from the beginning.

\item \textbf{Oracle Estimation Teaching.}
Although the model strictly follows the oracle’s step-by-step guidance, an incorrect identification of the peak in the third step leads to a significantly inaccurate final prediction.

\item \textbf{Few-Shot Examples.}
The ground-truth examples provided in the few-shot setting serve as effective calibration signals, leading to notably improved performance.

\item \textbf{Frame Index Provided.}
Providing frame indices helps the model better interpret the motion process. However, estimating the final value based solely on this information remains challenging.

\end{itemize}

\begin{figure*}[h]
    \centering
\includegraphics[width=\textwidth]{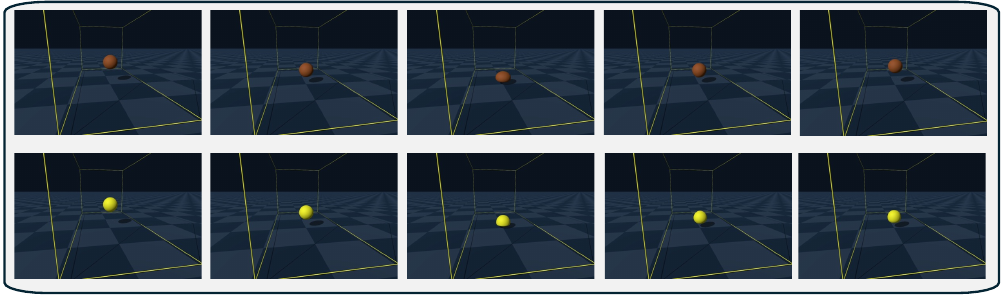}
\caption{
\textbf{Visual input for the \emph{relative formulation}.} The example is on Gemini for the elasticity property. Five representative frames are shown for the input videos. Top: the first video; Bottom: the second video. Ground truth should be $1$ in this case as the first video shows a higher elasticity value than the second one.
} 
\label{fig:mllm_example_rel}
\end{figure*}

\begin{figure*}[h]
    \centering
\includegraphics[width=0.9 \textwidth]{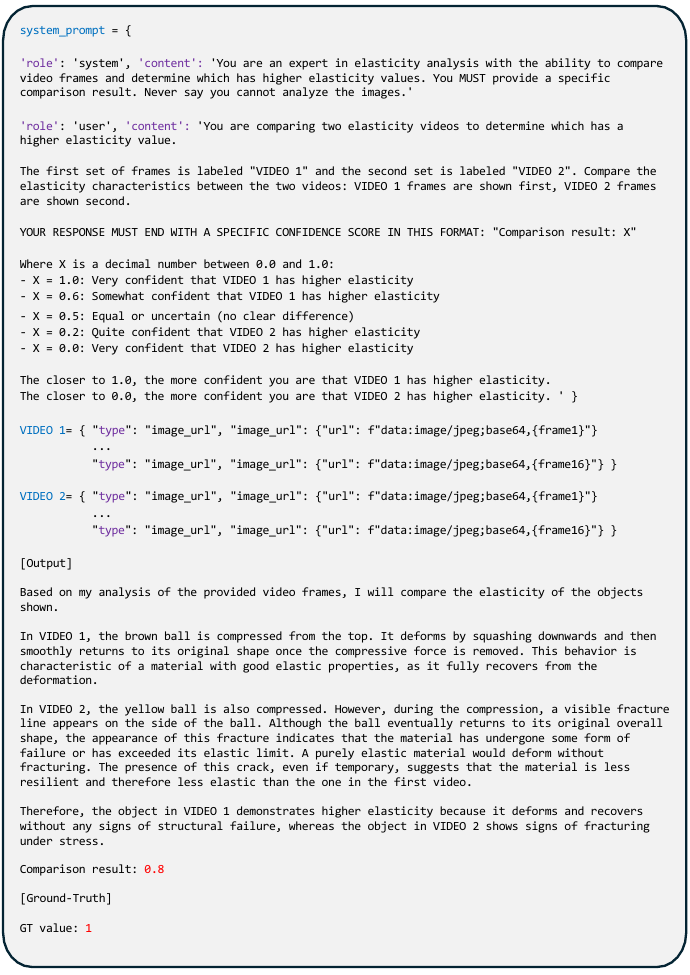}
\caption{
\textbf{Example of \emph{baseline prompt} for the \emph{relative formulation}.} The example is on Gemini for the elasticity property. The baseline model exhibits reasonable performance.
} 
\label{fig:mllm_example_rel_0}
\end{figure*}

\begin{figure*}[h]
    \centering
\includegraphics[width=0.55\textwidth]{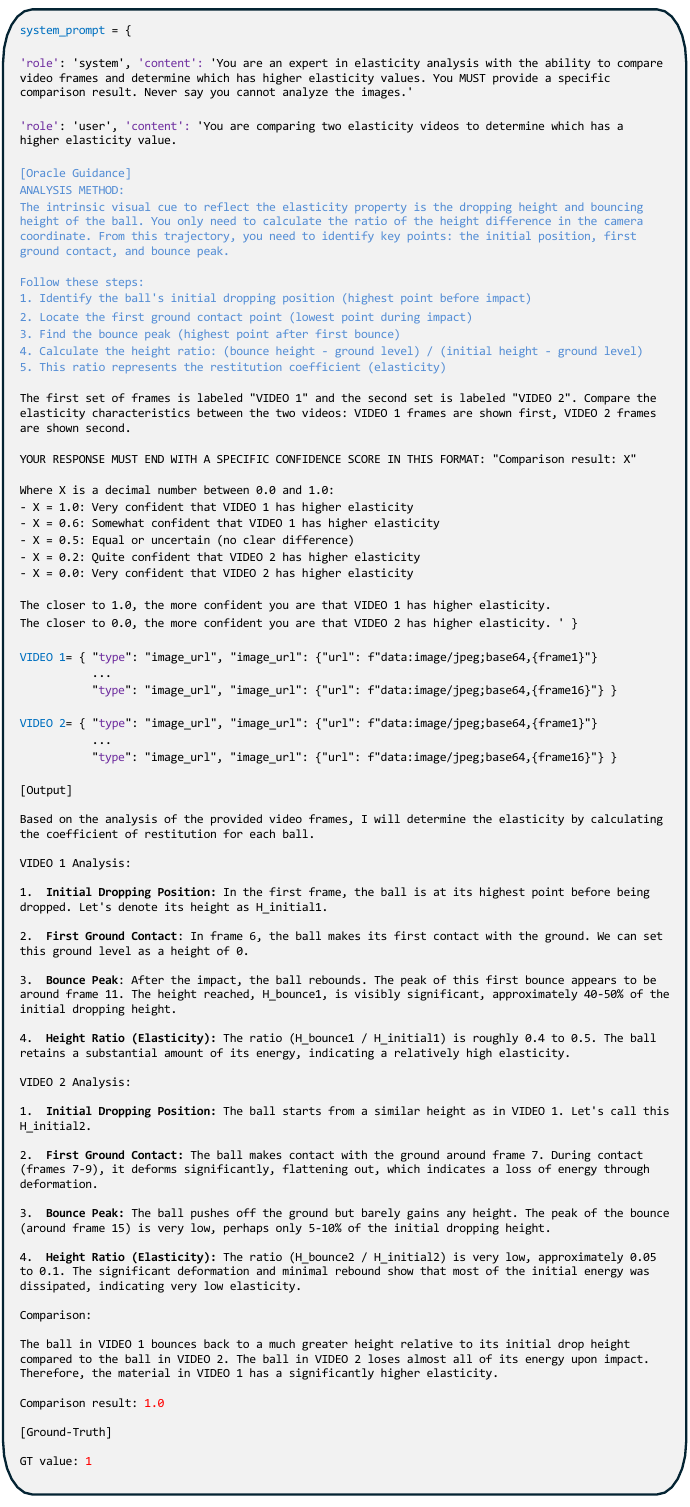}
\caption{
\textbf{Example of \emph{oracle estimation teaching} for the \emph{relative formulation}.} The example is on Gemini for the elasticity property. 
The oracle strategy promotes qualitative analysis ({\em e.g.}, comparing motion or relative magnitudes) without forcing exact calculations. This flexible reasoning process leads to more reliable outputs.
} 
\label{fig:mllm_example_rel_1}
\end{figure*}

\begin{figure*}[h]
    \centering
\includegraphics[width=0.7\textwidth]{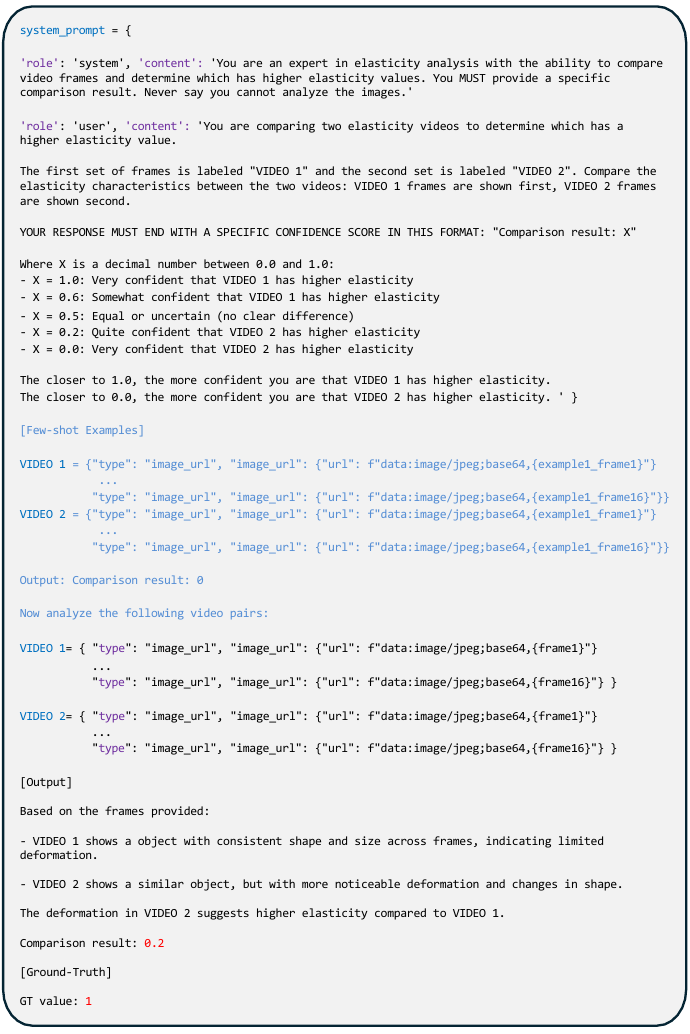}
\caption{
\textbf{Example of \emph{few-shot examples} for the \emph{relative formulation}.} The example is on Gemini for the elasticity property. 
The relative task is simpler—determining which of two instances has a greater physical value—without requiring exact numerical estimates. Here, few-shot examples tend to degrade performance, often encouraging shortcut responses that reduce interpretability and stability. 
} 
\label{fig:mllm_example_rel_2}
\end{figure*}

\begin{figure*}[h]
    \centering
\includegraphics[width=0.8 \textwidth]{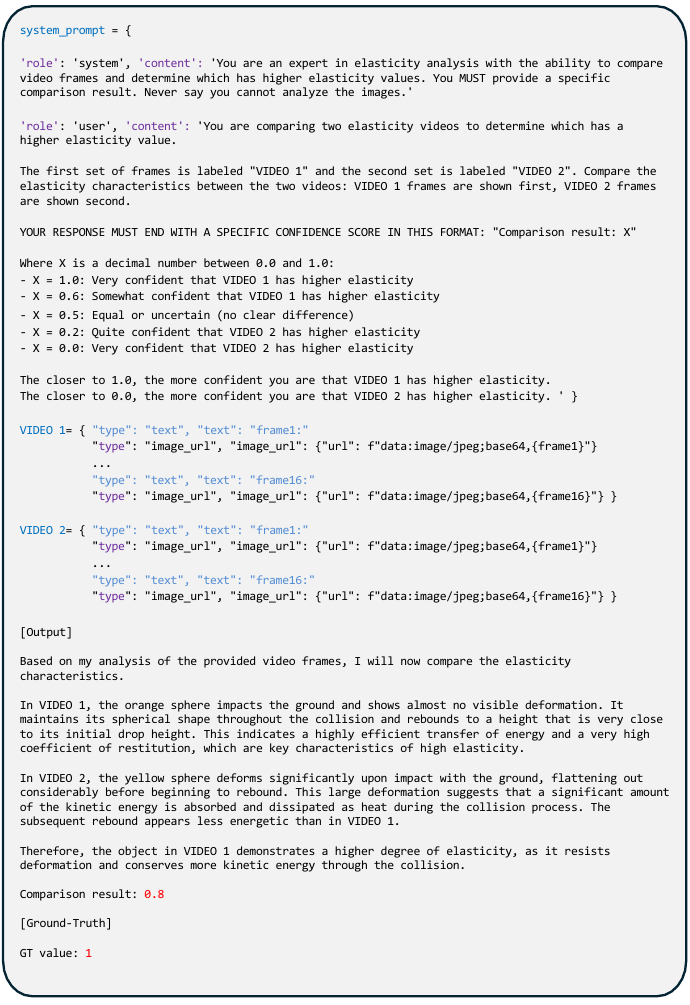}
\caption{
\textbf{Example of \emph{frame index provided} for the \emph{relative formulation}.} The example is on Gemini for the elasticity property. Providing the frame indices enhances the model's understanding of temporal dynamics, thereby resulting in more effective comparative reasoning.
} 
\label{fig:mllm_example_rel_3}
\end{figure*}

\begin{figure*}[h]
    \centering
\includegraphics[width=0.8 \textwidth]{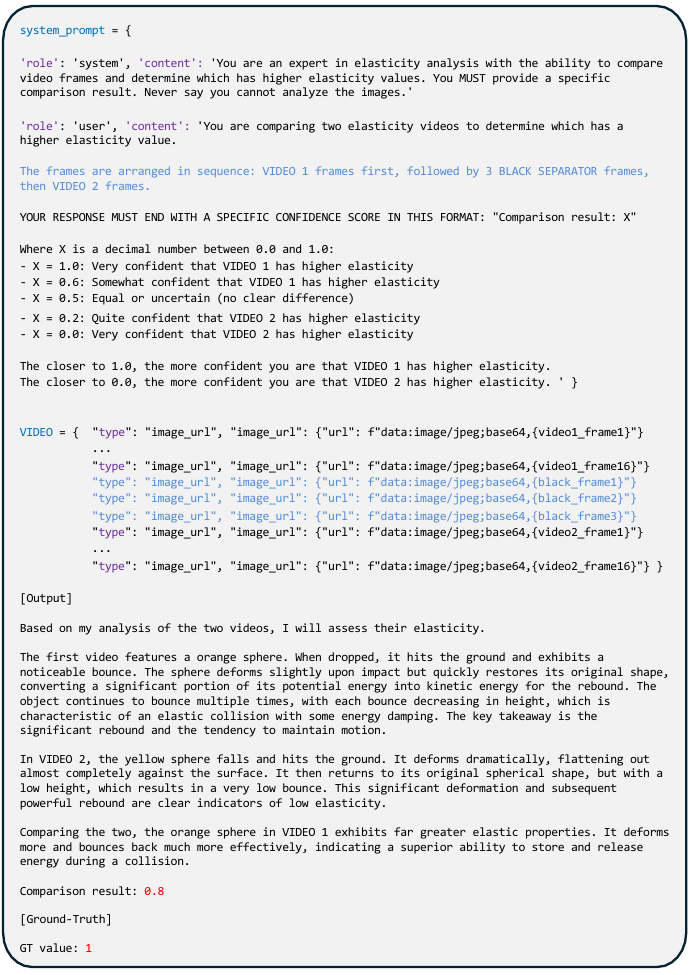}
\caption{
\textbf{Example of \emph{black frames in between} for the \emph{relative formulation}.} The example is on Gemini for the elasticity property. Concatenating both videos with black frames in between enables the model to better perform relative comparisons, likely by making inter-video relationships more explicit.
} 
\label{fig:mllm_example_rel_4}
\end{figure*}

\clearpage
The examples of \textbf{relative formulation} are provided in Figure~\ref{fig:mllm_example_rel} to Figure~\ref{fig:mllm_example_rel_4}. More specifically, Figure~\ref{fig:mllm_example_rel} shows the visual input to the MLLM; Figure~\ref{fig:mllm_example_rel_0} shows the prompt and model output of \emph{baseline prompt}; 
Figure~\ref{fig:mllm_example_rel_1} shows the prompt and model output of \emph{oracle estimation teaching}; 
Figure~\ref{fig:mllm_example_rel_2} shows the prompt and model output of \emph{few-shot examples}; 
Figure~\ref{fig:mllm_example_rel_3} shows the prompt and model output of \emph{frame index provided};
Figure~\ref{fig:mllm_example_rel_4} shows the prompt and model output of \emph{black frames in between}.

It can be observed that:
\begin{itemize}

\item \textbf{Baseline Prompt.}
The baseline model exhibits reasonable performance.

\item \textbf{Oracle Estimation Teaching.}
The oracle strategy promotes qualitative analysis ({\em e.g.}, comparing motion or relative magnitudes) without forcing exact calculations. This flexible reasoning process leads to more reliable outputs.

\item \textbf{Few-Shot Examples.}
The relative task is simpler—determining which of two instances has a greater physical value—without requiring exact numerical estimates. Here, few-shot examples tend to degrade performance, often encouraging shortcut responses that reduce interpretability and stability. 

\item \textbf{Frame Index Provided.}
Providing the frame indices enhances the model's understanding of temporal dynamics, thereby resulting in more effective comparative reasoning.

\item \textbf{Black Frames in Between.}
Concatenating both videos with black frames in between enables the model to better perform relative comparisons, likely by making inter-video relationships more explicit.

\end{itemize}

\clearpage
\section{Additional Scatter Plots}
\label{sec:additional_scatter_plot}
\vspace{5pt}

As mentioned in Section~\ref{sec:exp_qualitative} of the main paper, in this section we provide more scatter plots for different models on different test splits of the three dynamic physical properties.

As the evaluation for absolute prediction can automatically align the range of the predicted values with the ground truth values, in the scatter plot we show the raw prediction values without aligning them to the range of ground truth values.

\begin{figure*}[h]
    \centering
\includegraphics[width=0.9\textwidth]{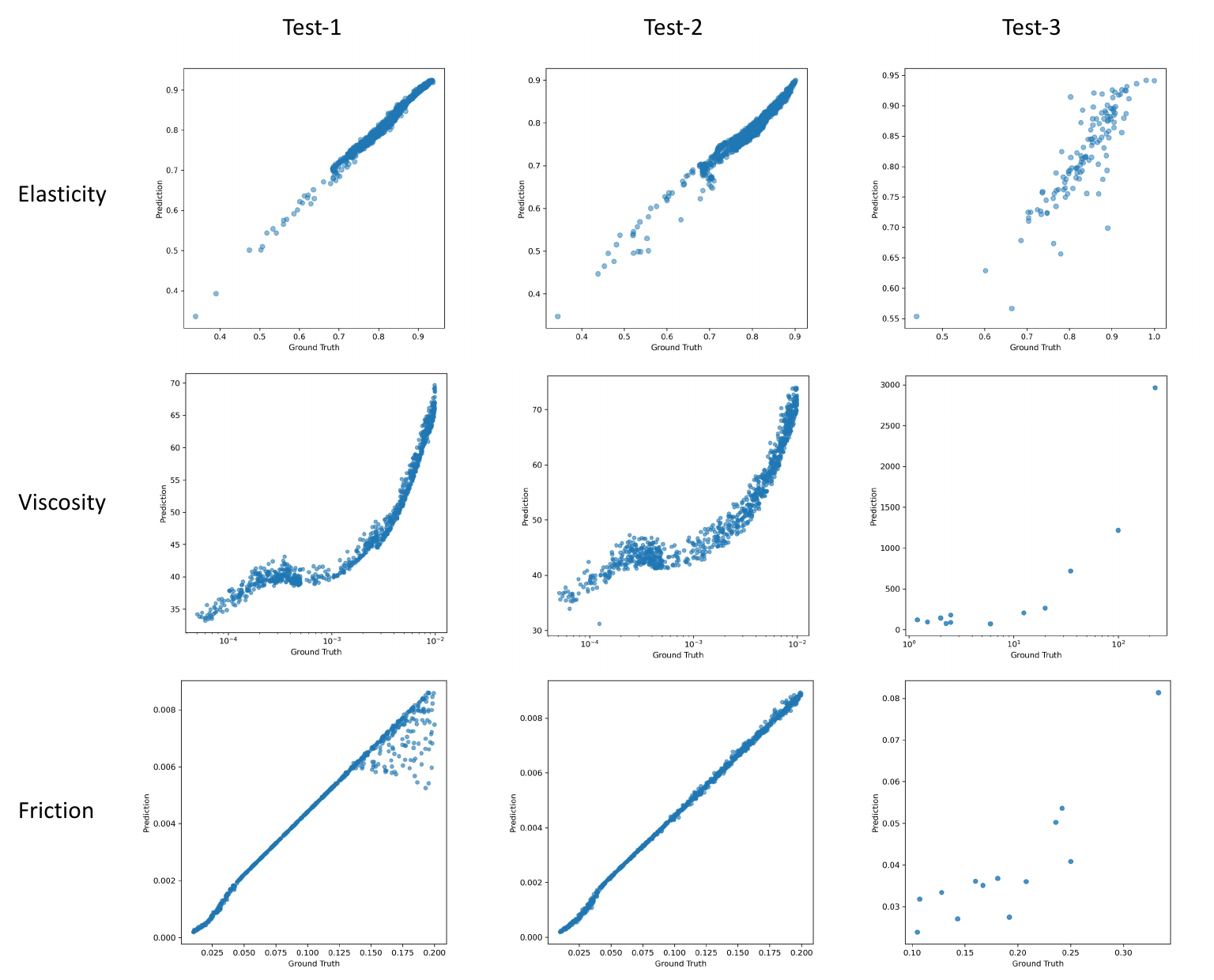}
\caption{
\textbf{Scatter plots for Oracle Estimation.} Top Row: Elasticity; Middle Row: Viscosity; Bottom Row: Friction. Left Column: Test-1; Middle Column: Test-2; Right Column: Test-3. For viscosity \texttt{test-3}, for each liquid, we take an average of the predictions for all samples to make the scatter plot, so that we can reduce the noise introduced by a single pouring liquid experiment; For friction \texttt{test-3}, for each combination of object and surface, we take an average of the predictions for all samples to make the scatter plot, so that we can reduce the noise introduced by a single sliding object experiment.
} 
\label{fig:scatter_oracle}
\end{figure*}

Figure~\ref{fig:scatter_oracle} shows the scatter plots of oracle estimation on different test splits of the three dynamic physical properties.

\begin{figure*}[h]
    \centering
\includegraphics[width=\textwidth]{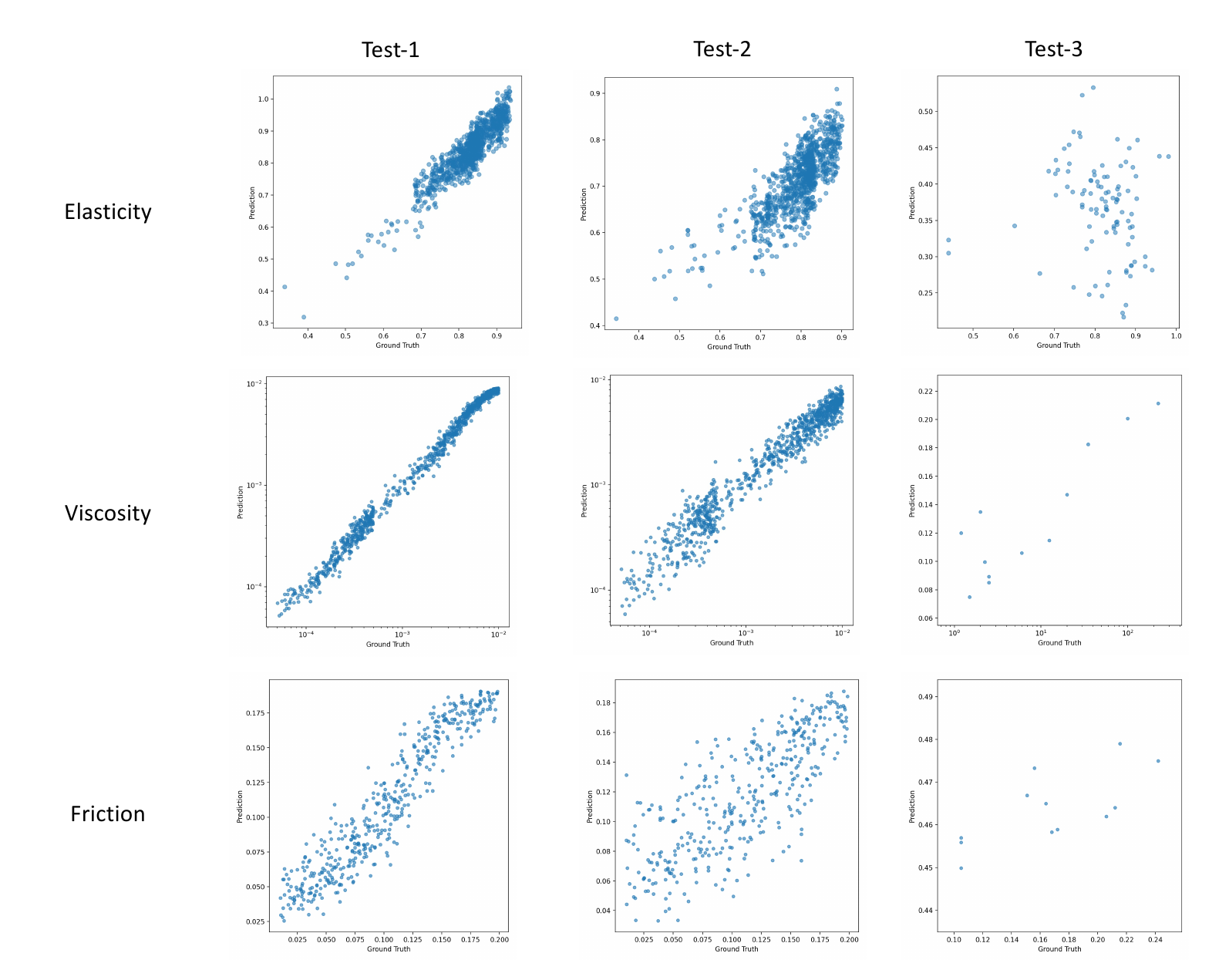}
\caption{
\textbf{Scatter plots for Video Generative Model.} Top Row: Elasticity; Middle Row: Viscosity; Bottom Row: Friction. Left Column: Test-1; Middle Column: Test-2; Right Column: Test-3. For viscosity \texttt{test-3}, for each liquid, we take an average of the predictions for all samples to make the scatter plot, so that we can reduce the noise introduced by a single pouring liquid experiment; For friction \texttt{test-3}, for each combination of object and surface, we take an average of the predictions for all samples to make the scatter plot, so that we can reduce the noise introduced by a single sliding object experiment.
} 
\label{fig:scatter_dynamicrafter}
\end{figure*}

\clearpage
Figure~\ref{fig:scatter_dynamicrafter} shows the scatter plots of DynamiCrafter on different test splits of the three dynamic physical properties. 

\begin{figure*}[h]
    \centering
\includegraphics[width=\textwidth]{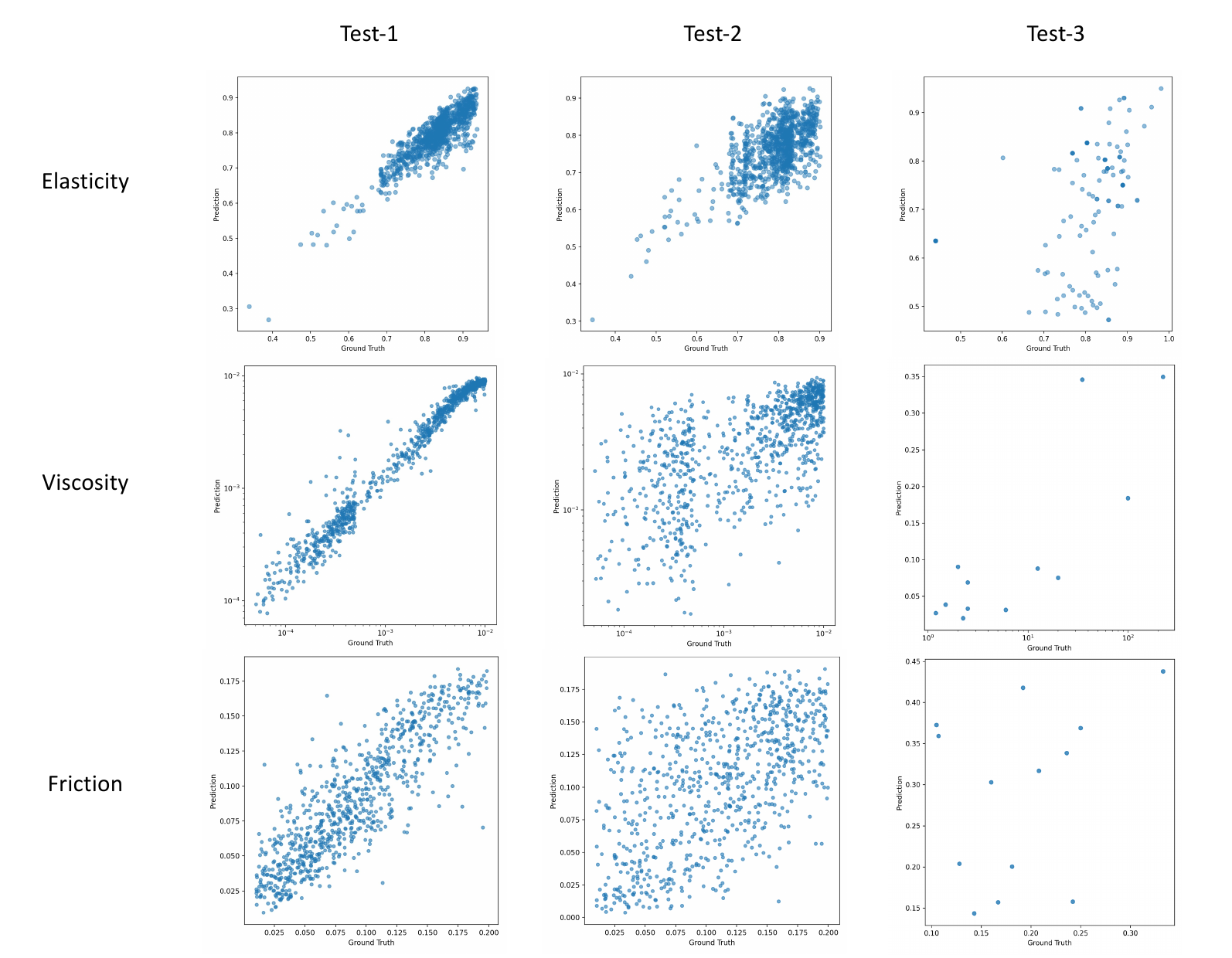}
\caption{
\textbf{Scatter plots for Video Self-Supervised Model.} Top Row: Elasticity; Middle Row: Viscosity; Bottom Row: Friction. Left Column: Test-1; Middle Column: Test-2; Right Column: Test-3. For viscosity \texttt{test-3}, for each liquid, we take an average of the predictions for all samples to make the scatter plot, so that we can reduce the noise introduced by a single pouring liquid experiment; For friction \texttt{test-3}, for each combination of object and surface, we take an average of the predictions for all samples to make the scatter plot, so that we can reduce the noise introduced by a single sliding object experiment.
} 
\label{fig:scatter_vjepa}
\end{figure*}

\clearpage
Figure~\ref{fig:scatter_vjepa} shows the scatter plots of V-JEPA-2 on different test splits of the three dynamic physical properties. 

\begin{figure*}[h]
    \centering
\includegraphics[width=\textwidth]{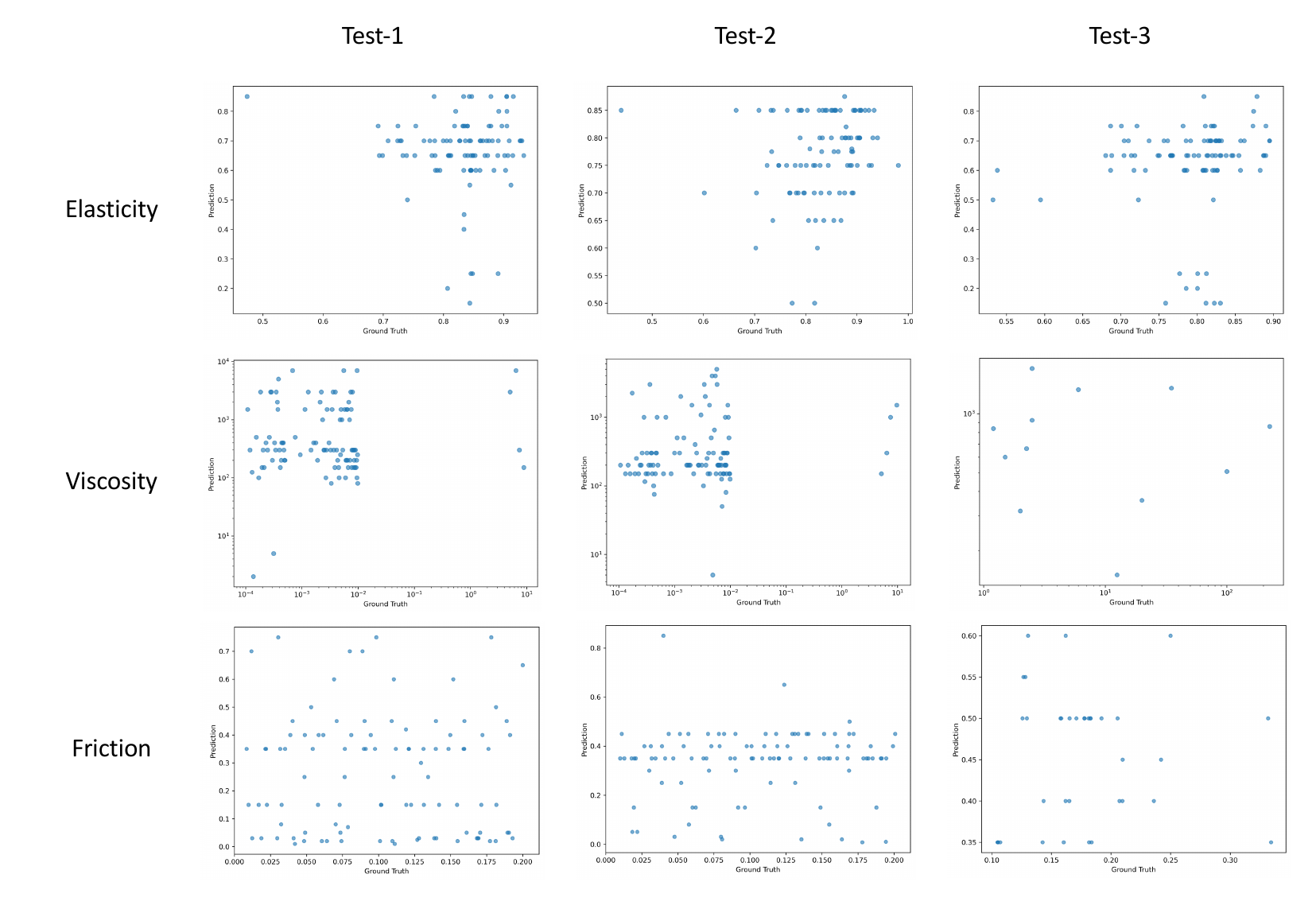}
\caption{
\textbf{Scatter plots for MLLMs (Qwen2.5VL-max).} Top Row: Elasticity; Middle Row: Viscosity; Bottom Row: Friction. Left Column: Test-1; Middle Column: Test-2; Right Column: Test-3. For viscosity \texttt{test-3}, for each liquid, we take an average of the predictions for all samples to make the scatter plot, so that we can reduce the noise introduced by a single pouring liquid experiment; For friction \texttt{test-3}, for each combination of object and surface, we take an average of the predictions for all samples to make the scatter plot, so that we can reduce the noise introduced by a single sliding object experiment.
} 
\label{fig:scatter_qwen}
\end{figure*}

\clearpage
Figure~\ref{fig:scatter_qwen} shows the scatter plots of Qwen2.5VL-max on different test splits of the three dynamic physical properties. For \texttt{test-1} and \texttt{test-2}, due to the limitation of resources, a random subset of 100 samples are used.

\begin{figure*}[h]
    \centering
\includegraphics[width=\textwidth]{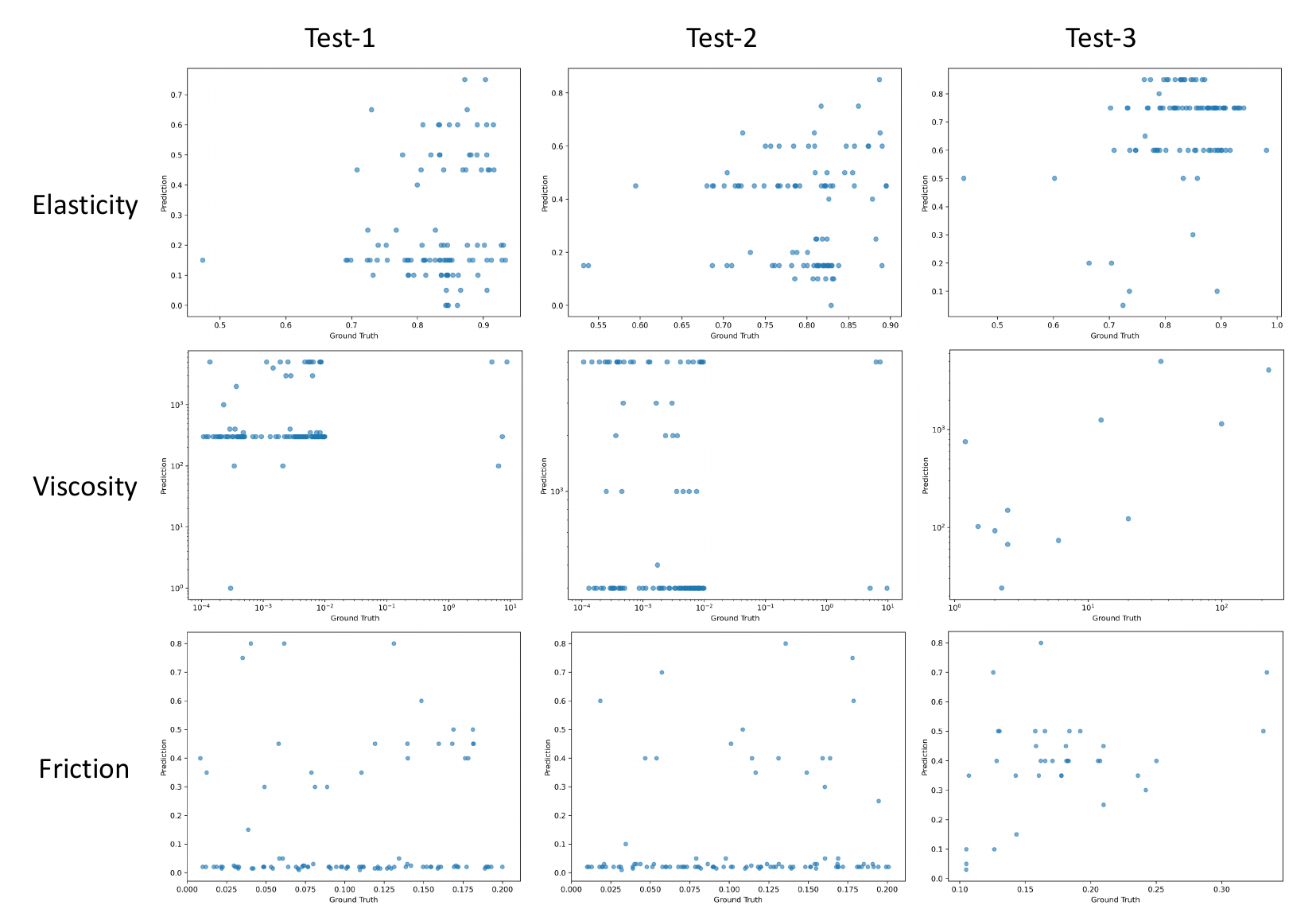}
\caption{
\textbf{Scatter plots for MLLMs (GPT-4o).} Top Row: Elasticity; Middle Row: Viscosity; Bottom Row: Friction. Left Column: Test-1; Middle Column: Test-2; Right Column: Test-3. For viscosity \texttt{test-3}, for each liquid, we take an average of the predictions for all samples to make the scatter plot, so that we can reduce the noise introduced by a single pouring liquid experiment; For friction \texttt{test-3}, for each combination of object and surface, we take an average of the predictions for all samples to make the scatter plot, so that we can reduce the noise introduced by a single sliding object experiment.
} 
\label{fig:scatter_gpt}
\end{figure*}

\clearpage
Figure~\ref{fig:scatter_gpt} shows the scatter plots of GPT-4o on different test splits of the three dynamic physical properties. For \texttt{test-1} and \texttt{test-2}, due to the limitation of resources, a random subset of 100 samples are used.

\begin{figure*}[h]
    \centering
\includegraphics[width=\textwidth]{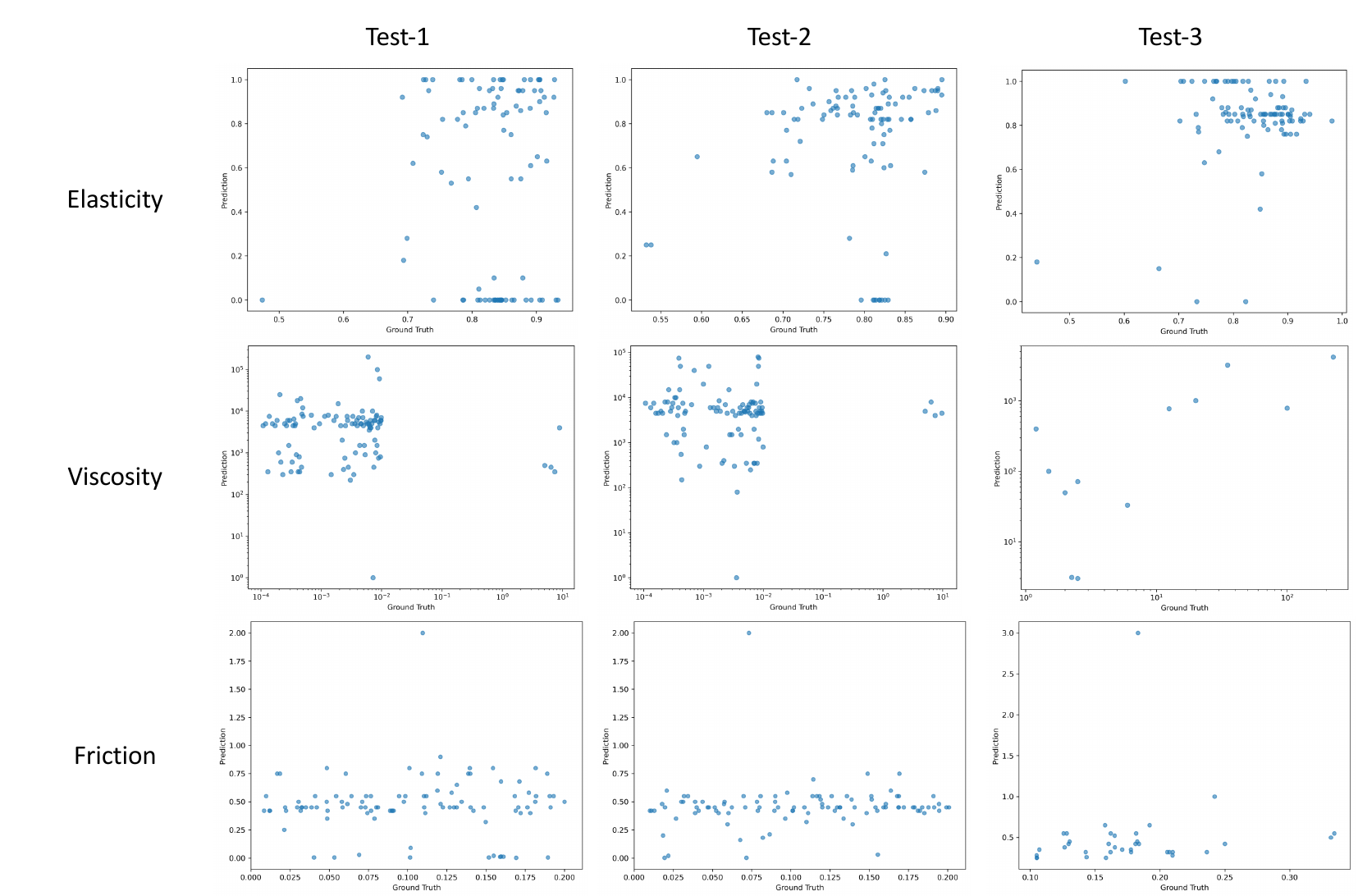}
\caption{
\textbf{Scatter plots for MLLMs (Gemini-2.5-pro).} Top Row: Elasticity; Middle Row: Viscosity; Bottom Row: Friction. Left Column: Test-1; Middle Column: Test-2; Right Column: Test-3. For viscosity \texttt{test-3}, for each liquid, we take an average of the predictions for all samples to make the scatter plot, so that we can reduce the noise introduced by a single pouring liquid experiment; For friction \texttt{test-3}, for each combination of object and surface, we take an average of the predictions for all samples to make the scatter plot, so that we can reduce the noise introduced by a single sliding object experiment.
} 
\label{fig:scatter_gemini}
\end{figure*}

\clearpage
Figure~\ref{fig:scatter_gemini} shows the scatter plots of Gemini-2.5-pro on different test splits of the three dynamic physical properties. For \texttt{test-1} and \texttt{test-2}, due to the limitation of resources, a random subset of 100 samples are used.

